\documentclass[twocolumn]{article}

\usepackage{ifluatex,ifxetex}
\ifluatex
\usepackage{fontspec} 
\else\ifxetex
\usepackage{fontspec} 
\else 
\usepackage[T1]{fontenc}
\usepackage[utf8]{inputenc} 
\fi\fi

\usepackage{array}
\usepackage{amsmath}
\usepackage{amssymb}
\usepackage{amsthm}
\usepackage[linesnumbered,ruled,vlined]{algorithm2e}

\SetCommentSty{mycommfont}
\usepackage{authblk}

\usepackage{bbm}
\usepackage{booktabs}

\usepackage{caption}
\usepackage{cite}
\usepackage{csvsimple}
\usepackage{color}
\usepackage{csvsimple}

\usepackage{diagbox}
\usepackage{dsfont}

\usepackage{graphicx}
\usepackage{geometry}

\usepackage[pagebackref=true,breaklinks=true,letterpaper=true,colorlinks,bookmarks=false]{hyperref}

\usepackage{multirow}
\usepackage{makecell}
\usepackage{multirow}
\usepackage{makecell}
\usepackage{multirow}
\usepackage{makecell}

\usepackage{paralist}
\usepackage{pgfplots}
\usepackage{pifont}      
\usepackage{subfigure}

\usepackage{tikz}

\usepackage{url}

\usepackage{xspace}

\tikzset{mycolor/.code=\xdef\mycolor{#1},my color/.style={mycolor=#1,color=#1},
my pattern color/.style={mycolor=#1,pattern color=#1}}

\usetikzlibrary{pgfplots.groupplots}

\catcode`\_=11\relax

\def\_email#1@#2\q_nil{%
\href{mailto:#1@#2}{{\emailfont #1\emailampersat #2}}
}
\newcommand\emailfont{\sffamily}
\newcommand\emailampersat{{\color{black}\small@}}
\catcode`\_=8\relax 

\newcommand{\PreserveBackslash}[1]{\let\temp=\\#1\let\\=\temp}
\newcolumntype{C}[1]{>{\PreserveBackslash\centering}p{#1}}

\def\A{{\boldsymbol A}}
\def\a{{\boldsymbol a}}

\def\I{{\boldsymbol I}}

\def\p{{\boldsymbol p}}

\def\q{{\boldsymbol q}}

\def\T{{\boldsymbol T}}

\def\U{{\boldsymbol U}}

\def\V{{\boldsymbol V}}
\def\v{{\boldsymbol v}}
\def\W{{\boldsymbol W}}

\def\x{{\boldsymbol x}}
\def\Y{{\boldsymbol Y}}
\def\y{{\boldsymbol y}}
\def\Z{{\boldsymbol Z}}
\def\z{{\boldsymbol z}}

\def\DM{{\mathcal D}}

\def\RM{{\mathcal R}}

\def\XM{{\mathcal X}}

\def\diag{\mathtt{diag}}
\def\svd{\mathtt{svd}}

\def\softmax{\mathtt{softmax}}
\def\mean{\mathtt{mean}}
\def\mod{\mathtt{mod}}

\def\argmax{\mathop{\rm argmax}}

\def\method{{MIE}}
\def\CREMAD{{CREMA-D}}
\def\Kinetics{{Kinetics-Sounds}}
\def\Twitter2015{{Twitter2015}}
\def\Sarcasm{{Sarcasm}}
\def\NVGesture{{NVGesture}}
\newgeometry{left=2cm,right=2cm,top=2cm,bottom=2cm}

\newcolumntype{C}[1]{>{\centering\let\newline\\\arraybackslash\hspace{0pt}}m{#1}}
\newcolumntype{L}[1]{>{\let\newline\\\arraybackslash\hspace{0pt}}m{#1}}

\font\emailfont=cmr12 at 9pt

\begin{document}
\title{\bf Multimodal Classification via Modal-Aware Interactive Enhancement}
\date{\today}
\author[ ]{Qing-Yuan Jiang}
\author[ ]{Zhouyang Chi}
\author[ ]{Yang Yang*}

\affil[ ]{Nanjing University of Science and Technology}
\affil[ ]{\texttt{qyjiang24@gmail.com}, 
\texttt{1527612210@njust.edu.cn},
\texttt{yyang@njust.edu.cn}}
\date{}
\maketitle

\begin{abstract}
Due to the notorious modality imbalance problem, multimodal learning~(MML) leads to the phenomenon of optimization imbalance, thus struggling to achieve satisfactory performance. Recently, some representative methods have been proposed to boost the performance, mainly focusing on adaptive adjusting the optimization of each modality to rebalance the learning speed of dominant and non-dominant modalities. To better facilitate the interaction of model information in multimodal learning, in this paper, we propose a novel multimodal learning method, called \underline{m}odal-aware \underline{i}nteractive \underline{e}nhancement~(\method). Specifically, we first utilize an optimization strategy based on sharpness aware minimization~(SAM) to smooth the learning objective during the forward phase. Then, with the help of the geometry property of SAM, we propose a gradient modification strategy to impose the influence between different modalities during the backward phase. Therefore, we can improve the generalization ability and alleviate the modality forgetting phenomenon simultaneously for multimodal learning. Extensive experiments on widely used datasets demonstrate that our proposed method can outperform various state-of-the-art baselines to achieve the best performance.
\end{abstract}
\section{Introduction}
\footnote{Corresponding author: Yang Yang.}
Human beings perceive the world through five senses~\cite{MML:journals/corr/abs-2301-04856,MML:journals/pami/SunKRBWL23}, including hearing, touch, smell, taste, and sight. Multimodal data collected from different sensors provides different perspectives on understanding the world. Inspired by multimodal data processing ability of humans, multimodal learning~(MML)~\cite{Film:conf/aaai/PerezSVDC18,MML:journals/pami/BaltrusaitisAM19,MML:journals/access/GuoWW19,OGR-GB:conf/cvpr/WangTF20,DOMFN:conf/mm/0074ZGGZ22,MSLR:conf/acl/YaoM22,Greedy:conf/icml/WuJCG22,ModalityLaziness:conf/iclr/ChengTLLWYZ22,CoHOZ:conf/mm/LiaoLLL00Y22,MGAD:conf/mm/HuangTDX22,MV-SSTMA:conf/mm/LiWZL22,AF:journals/pami/LiangQGCL22,OGM:conf/cvpr/PengWD0H22,MML:journals/pami/XuZC23,PMR:conf/cvpr/Fan0WW023,MLA:journals/corr/abs-2311-10707} has attracted much attention and made promising progress across a wide range of real applications including speech recognition~\cite{AVSR:journals/cm/YuhasGS89,MMDL:conf/icml/NgiamKKNLN11}, action classification~\cite{Classification:journals/mta/LanBYLH14}, image caption~\cite{Caption:conf/acl/ChangMSPM15,Detection:conf/ijcai/HodoshYH15,NIC:conf/aaai/FuSZY24}, multimedia retrieval~\cite{CMR:journals/corr/WangY0W016,CMR:journals/corr/abs-2308-14263,AEIR:journals/fcsc/YangGLLLY24}, recommendation system~\cite{GMMF:conf/mm/XiaoDCJYDL22,CKGE:journals/tois/YangZSDZL24,KA-MemNN:conf/wsdm/ZhuCLYYX20} and so on~\cite{TPCL:journals/corr/abs-2405-06926,CAT:conf/aaai/YangHGXX23,CMML:conf/ijcai/YangWZX019,DRUMN:conf/aaai/YangWZL019}.

\begin{figure}[t]
\centering
\includegraphics[width=0.85\linewidth]{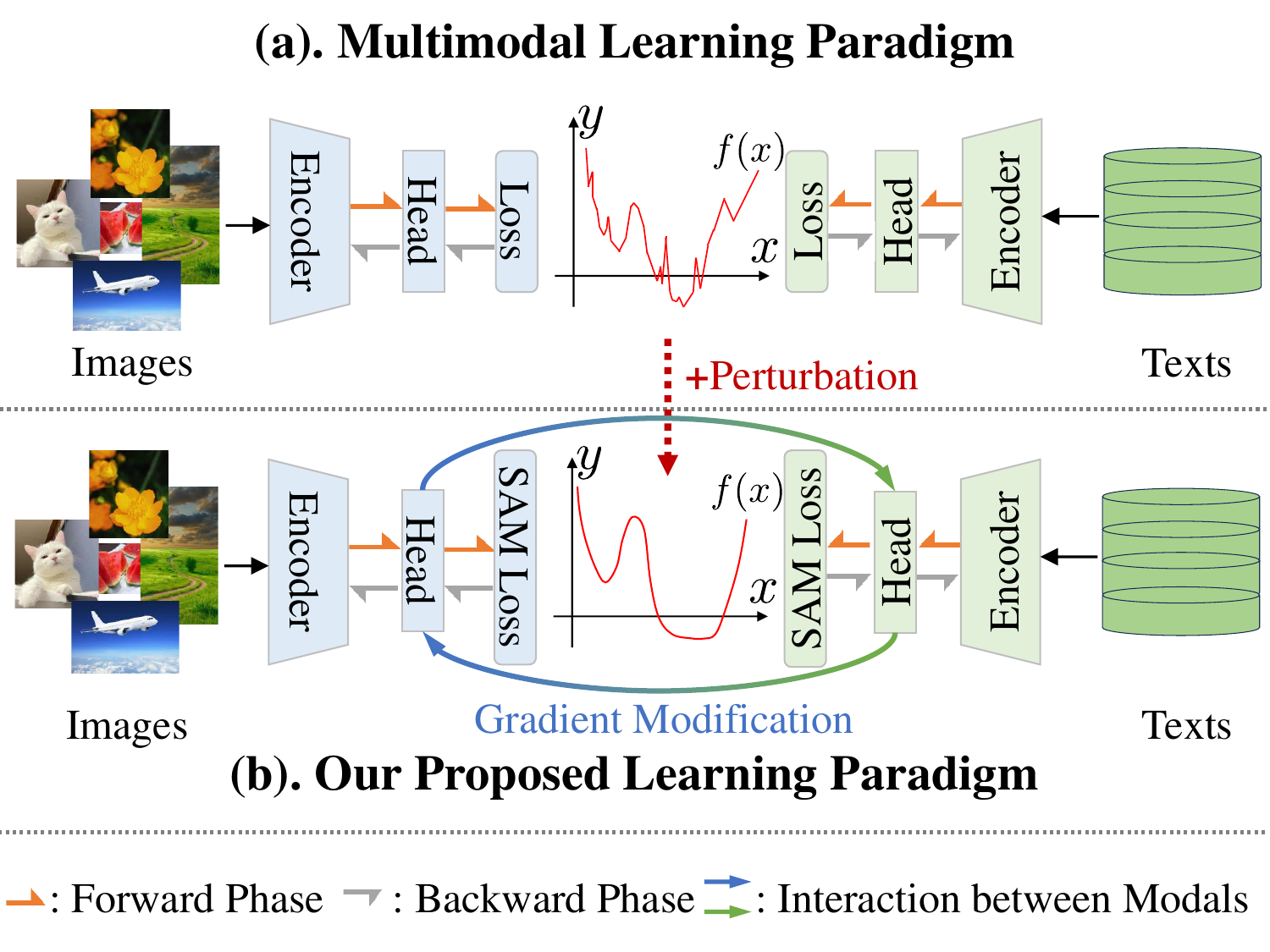}
\caption{Comparison of different paradigms. (a). Multimodal learning paradigm. (b). The proposed paradigm in this paper. Compared with traditional MML paradigm, we design a SAM-based optimization strategy by introducing a perturbation parameters. We further utilize a gradient modification strategy to enhance the interaction between different modalities.}
\label{fig:paradigm}
\end{figure}

Compared with the single modality method, multimodal learning methods are expected to achieve better performance through integrating rich information from multiple modalities. However, as the information between the different modalities is unbalanced, multimodal learning, which usually tries to optimize a uniform objective, falls into the trap of focusing on learning the dominant modality while ignoring the non-dominant modality~\cite{OGR-GB:conf/cvpr/WangTF20,MAN:journals/corr/abs-2011-06102,ATF:journals/spl/SunMH21,OGM:conf/cvpr/PengWD0H22}. Hence, the performance of multimodal learning in practical applications is greatly restricted because of the modality imbalance problem. Essentially, this is due to the different convergence rates~\cite{MAN:journals/corr/abs-2011-06102,ATF:journals/spl/SunMH21} of different modalities during training. 

\begin{figure*}[t]
\centering
\includegraphics[height=0.38\textwidth]{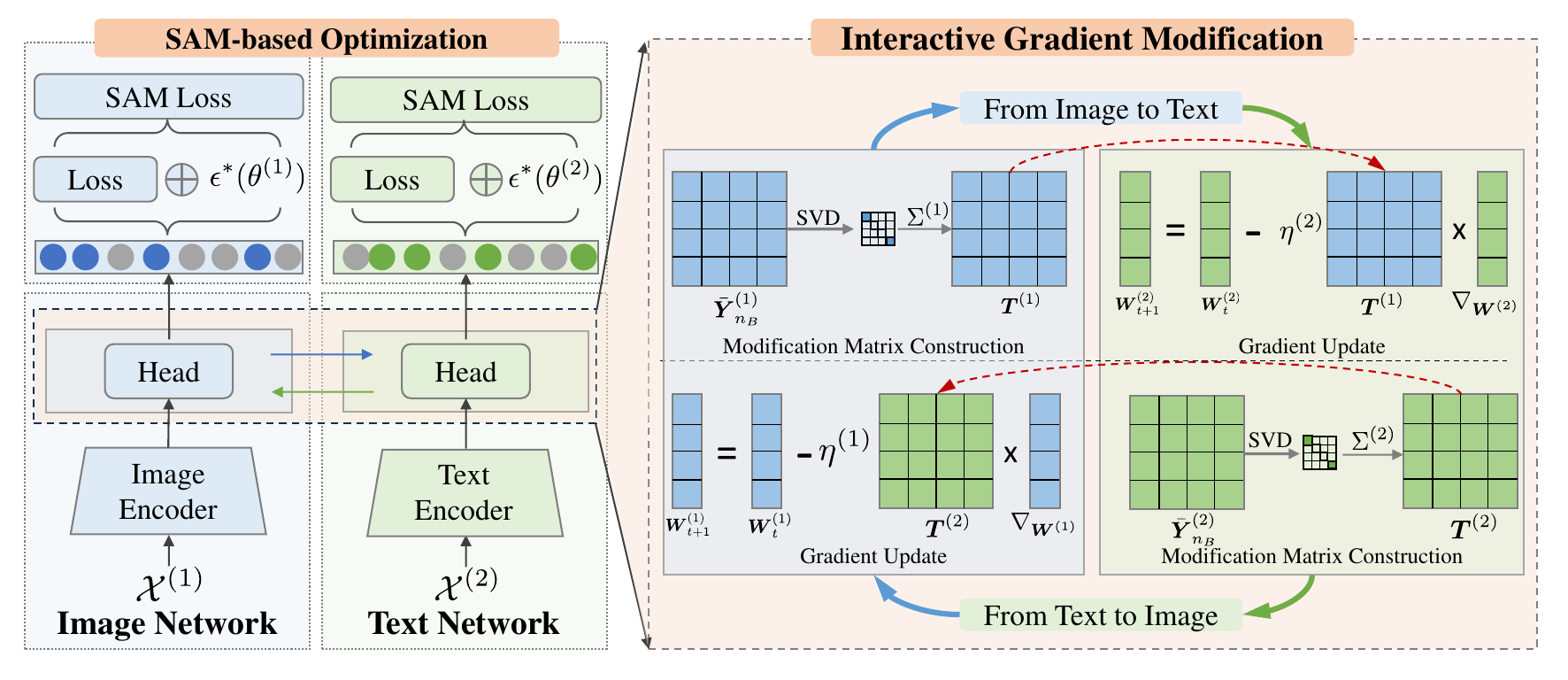} 
\caption{The architecture of our proposed~\method. Our method contains two key components, i.e., SAM-based optimization~(shown in the upper left corner of the panel) and interactive gradient modification~(shown in the right part of the panel).}
\label{fig:arch}
\end{figure*}

In recent years, many works~\cite{OGR-GB:conf/cvpr/WangTF20,nvGesture:conf/icml/WuJCG22,OGM:conf/cvpr/PengWD0H22,PMR:conf/cvpr/Fan0WW023} have explored the modality imbalance problem and various algorithms have been proposed to balance the learning procedure for all modalities. The general paradigm of these methods is illustrated in Figure~\ref{fig:paradigm}~(a). Among these methods, most of them~\cite{OGR-GB:conf/cvpr/WangTF20,OGM:conf/cvpr/PengWD0H22,PMR:conf/cvpr/Fan0WW023,AGM:conf/iccv/LiLHLLZ23} focus on designing a learning adjustment strategy to rebalance the learning speeds for different modalities. Other representative methods~\cite{UMT:journals/corr/abs-2106-11059,nvGesture:conf/icml/WuJCG22} introduce extra networks as the auxiliary module to overcome the modality imbalance problem. Unfortunately, these methods usually adopt a joint adaptive optimization strategy to learn relatively isolated encoders for each modality, thus neglecting the interaction between all modalities and encountering the modality forgetting problem inevitably. Recent work MLA~\cite{MLA:journals/corr/abs-2311-10707} designs an alternating uni-modal adaption algorithm to capture the cross-modality information. MLA enhances the interaction between different modalities to some extent and alleviates the modal forgetting problem. However, as the ubiquitous overparameterization phenomenon~\cite{SAM:conf/iclr/ForetKMN21} in deep learning, the generalization ability might be limited for multimodal learning. Hence, for existing multimodal learning methods including MLA, the alternating uni-modal learning process between different modalities might result in a sub-optimal solution.

How to improve the generalization ability of models in multimodal learning? Essentially, flatness and sharpness~\cite{Flatness:conf/iclr/ChaudhariCSLBBC17,Sharpness:conf/iclr/KeskarMNST17,SAM:conf/iclr/ForetKMN21} characterize the nature of the loss and have a significant impact on the generalization of the model. Sharpness aware minimization~(SAM)~\cite{SAM:conf/iclr/ForetKMN21} proposes to optimize loss value and loss sharpness simultaneously, leading to improving the generalization. Drawing support from the SAM, we can improve the generalization of a specific model to different modalities. Intuitively, improving the generalization ability of the model of different modalities can help reduce the difficulty of interactive learning between different modalities.

In this paper, to improve the generalization ability and overcome the modality forgetting problem,  we propose a novel multimodal learning method, called modal-aware interactive enhancement~(\method), in multimodal learning. Specifically, by introducing SAM into multimodal learning, we train the models by minimizing the loss value and loss sharpness during the forward phase. Furthermore, by using the geometry property of SAM-based optimization, we design a novel gradient modification strategy to impose the influence of a well-learned model on the model to be learned during the backward phase. The paradigm of our proposed method is shown in Figure~\ref{fig:paradigm}~(b). Our contributions can be summarized as follows: (1). We design an optimization process by introducing the SAM-based optimization for multimodal learning during the forward phase. Based on SAM, the generalization ability between different isolated modalities can be improved. (2). During the backward phase, we propose a novel gradient modification strategy with the help of the geometry property of SAM. By using this strategy, the modality forgetting phenomenon is alleviated. (3). Extensive experiments on widely used datasets show that our \method~can outperform state-of-the-art~(SOTA) baselines to achieve the best performance.

The rest of the paper is organized as follows. Section~\ref{sec:related-work} introduces the related work including MML and SAM. \method~is presented in Section~\ref{sec:method}. The experiments are shown in Section~\ref{sec:exp}. Finally, we conclude our work in Section~\ref{sec:conclusion}.

\section{Related Works}\label{sec:related-work}
\subsection{Multimodal Learning}
The goal of multimodal learning approaches is to improve performance by fusing information from diverse sensors. From the perspective of information integration, multimodal learning approaches can be divided into early fusion, late fusion, and hybrid fusion approaches~\cite{MML:journals/pami/BaltrusaitisAM19,MML:journals/vc/BayoudhKHM22}. 

Early fusion methods~\cite{OGR-GB:conf/cvpr/WangTF20,DOMFN:conf/mm/0074ZGGZ22,MLA:journals/corr/abs-2311-10707,Early:journals/inffus/ZengYMH24} try to integrate multiple features when they are extracted through fusion strategies like concatenating, adding, or weighting. Early fusion methods aim to exploit the correlation of the low level features between different modalities. Since early fusion tries to fuse low-level features, early fusion might encounter the problem of heterogeneous data in some cases. In contrast, late fusion methods~\cite{LF:conf/sspr/MorvantHA14,Late:journals/cin/YeLLGY18,MSLR:conf/acl/YaoM22,Late:journals/eswa/AlfaroContrerasVIC23,MLA:journals/corr/abs-2311-10707,Late:journals/inffus/LiuHXXZD24} focus on integrating the final prediction of each model. Late fusion methods can be further categorized into hard and soft late fusion, where the former integrates based on category decisions and the latter on scores. The integration approaches for late fusion include summation, averaging, majority voting, and so on. For late fusion, the model training for each modality is independent of each other. Hybrid fusion methods~\cite{MMTM:conf/cvpr/JozeSIK20,Greedy:conf/icml/WuJCG22,Hybrid:conf/aaai/ZhengTWH023} try to combine the properties of early and late fusion. One advantage of hybrid fusion is that it can amalgamate the superiority of early and late fusion.  

\subsection{Sharpness Aware Minimization}
With the development of deep neural networks, overparameterization becomes a common phenomenon and causes a serious overfitting problem. The training loss of deep neural networks is usually highly non-convex. Many efforts have been made to overcome this problem by using the properties of the loss landscape. Sharpness aware minimization~(SAM)~\cite{SAM:conf/iclr/ForetKMN21,Flattening:conf/nips/DengCHWH21,FlatMatch:conf/nips/Huang0Y0L23} proposes an effective algorithm to improve the generalization ability by using the relationship between loss sharpness and generalization. In particular, instead of learning the original objective, SAM aims to minimize the loss value and loss sharpness simultaneously. The learned parameters by SAM usually lie in the neighborhoods that have uniformly loss value of the original objective, leading to converging to flat minima. Therefore, the objective will converge to a flat minimum and the generalization ability can be improved. SAM has been applied in many application scenarios successfully. For example, FlatMatch~\cite{FlatMatch:conf/nips/Huang0Y0L23} extends SAM to semi-supervised learning by penalizing the cross-sharpness between the worst-case model and the original model.

\section{Methodology}\label{sec:method}
In this section, we present our proposed multimodal representation learning method~\method~ in detail. The whole~\method~approach is shown in Figure~\ref{fig:arch}. \method~contains two important components, i.e., SAM-based optimization and flat gradient modification. In concretely, we first introduce SAM-loss optimization to improve the generalization ability of the model for each modality. Then, by leveraging the geometry property of SAM optimization, we design a gradient modification strategy to impose the influence between different modalities during the backward phase.

\subsection{Preliminary}
In this paper, we use boldface lowercase letters like $\a$ to denote vectors and uppercase letters like $\A$ to denote matrices. The $i$-th element of vector $\a$ is denoted as $a_i$. The element in the $i$-th row and $j$-th column of matrix $\A$ is denoted as $A_{ij}$. $\A^\top$ denotes the transpose of matrix $\A$. The diagonal matrix is denoted as $\diag(\cdot)$. Furthermore, we use $\svd(\cdot)$ to denote the singular value decomposition for a given matrix. The symbol $\mod(\cdot, \cdot)$ denotes the modulo function and $\mod(a, b)$ returns the remainder after division of $a$ by $b$. The summary of the notation definition is presented in the supplementary materials.

Assume that we have $n$ data entities for training, each of which contains $m$ modalities. Without loss of generality, we use $\DM=\{\XM^{(j)}\}_{j=1}^m$ to denote the training set, where $\XM^{(j)}=\{\x^{(j)}_1,\cdots,\x^{(j)}_n\}$ denotes the data points of $j$-th modality and $\x^{(j)}_i$ denotes the $i$-th data point. In addition, we are also given a category label $\y_i\in\{0,1\}^{c}$ for each data point, where $c$ denotes the number of category labels. In general, the goal of multimodal learning is to use the training set $\DM$ to learn a model to predict category labels for unseen data points.

In multimodal learning, late fusion approaches usually fuse scores or decisions provided by different isolated branches for each modality to predict the final category labels. For deep multimodal learning methods~\cite{OGR-GB:conf/cvpr/WangTF20,OGM:conf/cvpr/PengWD0H22,AGM:conf/iccv/LiLHLLZ23}, different deep neural networks are used as the model to predict categories for each modality. For the sake of simplicity, we use $\varphi^{(j)}(\cdot)$ to denote the encoder which is used to extract the feature of $j$-th modality. And the feature $\z^{(j)}_i$ of $i$-th data point can be calculated by $\z^{(j)}_i=\varphi^{(j)}(\x^{(j)}_i)$. Then the $j$-th prediction $\p_i^{(j)}\in\RM^c$ can be presented as:
\begin{align}
\p_i^{(j)}=\phi^{(j)}(\x^{(j)}_i;\theta^{(j)})=\softmax([\W^{(j)}]^\top\z^{(j)}_i),\nonumber
\end{align}
where $\W^{(j)}\in\RM^{d\times c}$ and $\theta^{(j)}$ denote the parameter of last full-connected layer and $j$-th classifier, respectively, and $\phi^{(j)}(\cdot)$ indicates the $j$-th classifier. According to $\p_i^{(j)}$, the training procedure is performed by minimizing the following loss function:
\small{\begin{align}
L(\theta^{(j)};\XM^{(j)})
&=\frac{1}{n}\sum_{i=1}^n\ell(\p_i^{(j)},\y_i;\theta^{(j)})=-\frac{1}{n}\sum_{i=1}^n\log\big([\y^{(j)}_i]^\top\p_i^{(j)}\big).\label{eq:CELoss}
\end{align}}

After training, the final prediction of $i$-th data point can be generated by the following equation:
\begin{align}
\p_i=f(\p^{(1)}_i,\cdots,\p_i^{(m)}).\nonumber
\end{align}
Here, $f(\cdot)$ denotes the late fusion strategy. In practice, there exist various late fusion strategies like averaging, maximization, and weighting. However, how to design fusion strategies is not the focus of our paper and it will be left for future study.

For commonly used late fusion multimodal approaches, the training procedure is performed independently on each modality. Most existing multimodal learning methods fail to integrate the overall modality information during isolated training procedures because of the notorious modality laziness problem~\cite{ModalityLaziness:conf/iclr/ChengTLLWYZ22,MLA:journals/corr/abs-2311-10707}. MLA~\cite{MLA:journals/corr/abs-2311-10707} designs a shared head structure to capture cross-modal interaction information. However, MLA ignores the overparameterization problem and fails to find a flat loss landscape while integrating information of isolated modality.

\subsection{\small{Modal-Aware Interactive
Enhancement}}
To track the problem of inability to interact with modality information during the isolated modality training procedure, we first introduce SAM~\cite{SAM:conf/iclr/ForetKMN21} into multimodal learning to improve the generalization ability of the model trained on one modality during the forward phase. Then we modify the gradient direction by using flat direction modification from another learned modality to mitigate the modality forgetting problem during the backward phase. In this way, we can enhance the feedback between different isolated modalities during both the forward and backward phases.

We use the training procedure of $k$-th and $l$-th modality to illustrate our proposed modal-aware interactive learning procedure. During the training procedure, we minimize the same objective when we train models for $k$-th and $l$-th modalities. Motivated by SAM, we try to learn parameters $\theta^{(k)}$ that lie in neighborhoods of $\theta^{(k)}$ having uniformly low loss. Specifically, we define the perturbation of parameter $\theta^{(k)}$ as $\epsilon$. Based on $\epsilon$, the SAM objective~\cite{SAM:conf/iclr/ForetKMN21,FlatMatch:conf/nips/Huang0Y0L23} can be defined as:
\begin{align}
L^{\text{SAM}}(\theta^{(k)})&\triangleq\max_{\epsilon:\Vert\epsilon\Vert_p\leq \rho}\;L(\theta^{(k)}+\epsilon),\nonumber\\
&\triangleq\max_{\epsilon:\Vert\epsilon\Vert_p\leq \rho}\;\frac{1}{n}\sum_{i=1}^{n}\ell(\theta^{(k)}+\epsilon;\p^{(k)}_i,\y_i),\nonumber
\end{align}
where $\rho$ restricts the perturbation magnitude of $\theta^{(k)}$ within $\ell_p$-ball. Instead of minimizing the objective function $L(\theta^{(k)})$ in Equation~(\ref{eq:CELoss}), we perturb the parameter $\theta^{(k)}$ with $\epsilon\in\Theta$ and optimize the following SAM objective:
\begin{align}
\min_{\theta^{(k)}}\;L^{\text{SAM}}(\theta^{(k)}).\label{eq:samloss}
\end{align}
Here, $\Theta$ denotes the parameter space. Through optimizing objective $L^{\text{SAM}}(\theta^{(k)})$, we can find a flatter loss landscape, thus improving the generalization ability.

In order to estimate the optimal perturbation $\epsilon^*$, we can construct the following inner maximization problem~\cite{SAM:conf/iclr/ForetKMN21}:
\begin{align}
\epsilon^*(\theta^{(k)})&=\argmax_{\Vert\epsilon\Vert_p\leq \rho}\;L(\theta^{(k)}+\epsilon)\nonumber\\
&\approx\argmax_{\Vert \epsilon\Vert_p\leq\rho}\;\epsilon^\top \nabla_{\theta^{(k)}}L(\theta^{(k)})\stackrel{p = 2}{\approx}\rho\frac{\nabla_{\theta^{(k)}}L(\theta^{(k)})}{\Vert\nabla_{\theta^{(k)}}L(\theta^{(k)})\Vert_2}.\label{eq:epsilon}
\end{align}

By substituting Equation~(\ref{eq:epsilon}) into SAM objective in Equation~(\ref{eq:samloss}) and differentiating, we can get:
\begin{align}
\nabla_{\theta^{(k)}} L^{\text{SAM}}&= \nabla_{\theta^{(k)}}\left[L\left(\theta^{(k)}+\epsilon^{*}(\theta^{(k)})\right)-L(\theta^{(k)})\right]+L(\theta^{(k)}) \nonumber\\ 
&\approx \nabla_{\theta^{(k)}} L\left(\theta^{(k)}+\epsilon^{*}(\theta^{(k)})\right)\nonumber \\
&=\frac{d\left(\theta^{(k)}+\epsilon^{*}(\theta^{(k)})\right)}{d \theta^{(k)}} \nabla_{\theta^{(k)}} L(\theta^{(k)})|_{\theta^{(k)}+\epsilon^{*}(\theta^{(k)})} \nonumber\\ 
&= \left.\nabla_{\theta^{(k)}} L(\theta^{(k)})\right|_{\theta^{(k)}+\epsilon^{*}(\theta^{(k)})}+o(\theta^{(k)}),\label{eq:SAM-gradient}
\end{align}
where $o(\theta^{(k)})$ denotes the second-order term with respect to $\theta^{(k)}$ and this term can be discarded to accelerate the computation. Intuitively, optimizing SAM objective can yield flatter minima which can generalize better compared with minimizing $L(\theta^{(k)})$ during the forward phase.

SAM-based optimization has a pivotal geometry property~\cite{SAM:conf/iclr/ForetKMN21}, i.e., the change of loss value is relatively smooth in the flat directions as the SAM objective has been well-learned to a flat loss landscape. Hence, to track the modality forgetting problem, we can design a strategy to modify the updating direction by using the flat directions of the well-learned $k$-th model when we train $l$-th modality, thus imposing the influence from well-learned $k$-th modality on $l$-th modality. Inspired by Adam-NSCL~\cite{Adam-NSCL:conf/cvpr/WangL0X21}, we design a singular value decomposition~(SVD) based approach to find the flat directions.

\begin{algorithm}[t]
\SetKwInOut{Input}{Input}
\SetKwInOut{Output}{Output}
\DontPrintSemicolon
\SetAlgoLined
\SetNoFillComment
\LinesNotNumbered 
\Input{~Training set $\DM=\{\XM^{(j)}\}^m_{j=1}$, labels $\Y=\{\y_i\}_{i=1}^n$.}
\Output{~The learned parameters $\{\theta^{(j)}\}_{j=1}^{(m)}$ of all models.}
\textbf{INIT} initialize gradient modification matrix. Initialize gradient modification matrix: $\forall k\in\{1,\cdots,m\},\;\T^{(k)}=\I$.\;
\For{$i=1\rightarrow \text{Out\_Iters}$}{
\tcc{iterate over all modalities.}
\For{$j=1\rightarrow m$}{
  \tcc{iterate over all training examples of $\XM^{(j)}$.}
  \For{$t=1\rightarrow \text{Inner\_Iters}$}{    
      Randomly construct a mini-batch $\XM^{(j)}_{t}$.\;
      Calculate loss $L(\theta^{(j)})$ for data point in $\XM^{(j)}_{t}$.\;
      Calculate $\epsilon^{*}(\theta^{(k)})$ according to Eq.~(\ref{eq:epsilon}).\;
      Calculate the gradient approximation $\nabla_{\theta^{(j)}}L^{\text{SAM}}$ of SAM loss according to Eq.~(\ref{eq:SAM-gradient}).\;
      Calculate modality index: $k=\mod(j+m-2, m)+1$.\;
      Update $\theta^{(j)}$: $\theta^{(j)}_{t+1}=\theta^{(j)}_t-\eta^{(j)}\T^{(k)}\nabla_{\theta^{(j)}}L^{\text{SAM}}$.\;            
  }
  \tcc{update all gradient modification matrix.}
  \For{$j=1\rightarrow n_B$}{
  Update cumulative variance according to Eq.~(\ref{eq:update-cov}).\;
  }
  Update $\T^{(j)}$ according to Eq.~(\ref{eq:update-T}).\;
}

}
\caption{Multimodal learning algorithm for \method}
\label{algo:MAIE}
\end{algorithm}

We utilize the full-connected layer before the classification layer to illustrate the flat direction modification strategy. Given $t$-th batch of samples $\XM^{(k)}_t=\{\x^{(k)}_1,\cdots,\x^{(k)}_{B}\}$, the features of the input batch can be calculated by:
\begin{align}
\Z_t^{(k)}=\varphi^{(k)}(\XM^{(k)}_t),\nonumber
\end{align}
where $\Z_t^{(k)}\in\RM^{B\times d}$, $d$ is the dimensionality of feature. Then we compute the mean of features and the covariance of the batch by:
\begin{align}
&\bar\z^{(k)}_t=\mean(\Z^{(k)}_t)\in\RM^{d},\nonumber\\
&\Y^{(k)}_t=\bar\z^{(k)}_t[\bar\z^{(k)}_t]^\top\in\RM^{d\times d}.\nonumber
\end{align}

Then, the cumulative variance can be calculated by:
\begin{align}
\left\{
\begin{aligned}
&\bar\Y^{(k)}_t=\Y^{(k)}_t,\;&&{\text{if }t=1,}\\
&\bar\Y^{(k)}_{t+1}=\bar\Y^{(k)}_{t}+\Y^{(k)}_t,\;&&{\text{otherwise}}.
\end{aligned}
\right.\label{eq:update-cov}
\end{align}

By applying SVD to $\bar\Y^{(k)}_{n_B}$, we have:
\begin{align}
\U^{(k)}\Lambda^{(k)}[\V^{(k)}]^\top\triangleq\svd(\bar\Y^{(k)}_{n_B}),\nonumber
\end{align}
where $n_B$ denotes the number of batch, $\Lambda=\diag(\lambda^{(k)}_1, \cdots, \lambda^{(k)}_q)$ and $\V^{(k)}=[\v^{(k)}_1,\cdots,\v^{(k)}_q]$ denote the singular values matrix and singular vectors, respectively.

For now, let us consider the geometry properties of the direction indicated  by singular vector $\v^{(k)}_i$. If we perturb $\Z_t^{(k)}$ along with the singular direction $\v^{(k)}_i$ with the perturbation magnitude $\gamma\v^{(k)}_i$, the change of the output for the last full-connected layer can be computed by:
\begin{align}
\Vert\Z_t^{(k)}\gamma\v^{(k)}_i\Vert=\Vert\U^{(k)}\Lambda^{(k)}[\V^{(k)}]^\top\gamma\v^{(k)}_i\Vert=\gamma\lambda^{(k)}_i.\label{eq:change}
\end{align}

From Equation~(\ref{eq:change}), the flatness of the direction indicated  by singular vector $\v^{(k)}_i$ is determined by the singular value $\lambda^{(k)}_i$. In other words, the larger the singular value, the smaller the update modification should be in the direction of the singular vector. Thus, we design the following gradient modification matrix $\T^{(k)}$:
\begin{align}
\T^{(k)}=\V^{(k)}\Sigma^{(k)}[\V^{(k)}]^\top.\label{eq:update-T}
\end{align}
Here, $\Sigma^{(k)}=\exp\big(-\frac{\tau}{\lambda^{(k)}_{\text{max}}-\lambda^{(k)}_{\text{min}}}({\Lambda^{(k)}-\lambda^{(k)}_{\text{min}}\I})\big)$, and $\tau>0$ is a scaling factor. Hence, when we update the parameter of $l$-th modality for the last full-connected layer, the SGD-based update rule is modified as:
\begin{align}
\W^{(l)}_{t+1}=\W^{(l)}_{t}-\eta^{(l)}\T^{(k)}\nabla_{\W^{(l)}}L^{\text{SAM}}(\theta^{(l)}),\label{eq:update-rule}
\end{align}
where $\eta^{(l)}$ denotes the corresponding learning rate. From Equation~(\ref{eq:update-rule}), we can find that the information of $k$-th modality is injected into $l$-th modality. Equipped with the flat direction modification strategy, the modality forgetting problem is greatly alleviated during the backward phase.

By introducing SAM-based optimization and gradient modification strategy, the interaction between each modality is greatly enhanced, thus leading to the generalization improvement and the alleviation of the modality forgetting problem. The learning algorithm of \method~is summarized in Algorithm~\ref{algo:MAIE}. 

Note that the aforementioned discussion is based on the assumption that the architecture of models of different modalities is the same. In scenarios where network architectures of different modalities are heterogeneous, the gradient modification strategy can be applied to deep layers of networks with the same architecture.

\subsection{Model Inference}
After training \method, the learned deep neural networks can be applied to predict for given unseen sample. Specifically, for a given sample that contains multiple modalities, the prediction of each modality can be generated by the corresponding model. Then, the late fusion strategy is adopted to generate the final prediction by fusing the predictions of each model.

\section{Experiments}\label{sec:exp}
\subsection{Datasets}
We adopt five widely used datasets, i.e., \CREMAD~\cite{CREMAD:journals/taffco/CaoCKGNV14}, \Kinetics~\cite{Kinetics-Sound:conf/iccv/ArandjelovicZ17}, \Twitter2015~\cite{Twitter15:conf/ijcai/Yu019}, \Sarcasm~\cite{Sarcasm:conf/acl/CaiCW19}, and \NVGesture~\cite{NVGeasture:conf/cvpr/MolchanovYGKTK16}, for evaluation. \CREMAD~and \Kinetics~datasets, which consist of audio and video modalities, are used for speech emotion recognition and video action recognition tasks, respectively. \CREMAD~dataset consists of 7,442 clips of 2$\sim$3 seconds from 91 actors. The clips are divided into 6,698 samples for training and 744 samples for testing. \Kinetics~dataset, which comprises 31 human action category labels, consists of 19,000 10-second clips. It is divided into a training set with 15K samples, a validation set with 1.9K samples, and a testing set with 1.9K samples. \Twitter2015~and \Sarcasm~datasets are collected from Twitter and consist of image and text modalities. \Twitter2015~contains 5,338 image-text pairs with 3,179 for training, 1,122 for validation, and 1,037 for testing. This dataset is used for emotion recognition. \Sarcasm~dataset is used for sarcasm detection task and consists of 24,635 image-text pairs. We split this dataset as 19,816 for training, 2,410 for validation, and 2,409 for testing following the setting of the original paper. \NVGesture~dataset contains 1,532 dynamic hand gestures. This dataset is divided into 1,050 for training and 482 for testing. To verify the effectiveness of our approach in scenarios with more than two modalities, we use RGB, Depth, and optical flow~(OF) modalities to carry out experiments for \NVGesture~dataset.

\begin{table*}[t]
\centering
\caption{Comparison with SOTA multimodal learning baselines.}  
\label{tab:main-exp}
\begin{tabular}{l|c|cc|cc|cc|cc}
\Xcline{1-10}{0.7pt}
{\multirow{2}{*}{Method}} & {\multirow{2}{*}{Fusion}} & \multicolumn{2}{c|}{{\CREMAD}}& \multicolumn{2}{c|}{{\Kinetics}}& \multicolumn{2}{c|}{{\Twitter2015}}& \multicolumn{2}{c}{{\Sarcasm}}\\\cline{3-10}
    &        & Accuracy& MAP     & Accuracy& MAP     & Accuracy& Mac-F1   & Accuracy & Mac-F1   \\
\Xcline{1-10}{0.7pt}
Greedy & Hybrid & 66.64\% & 72.64\% & 66.52\% & 72.81\% & -       & -        & -        & -        \\\hline
OGR-GB & Concat & 64.65\% & 68.54\% & 67.10\% & 71.39\% & 74.35\% & 68.69\%  & 83.35\%	& 82.71\%  \\
OGM    & Concat & 66.94\% & 71.73\% & 66.06\% & 71.44\% & \underline{74.92\%} & 68.74\%  & 83.23\%  & 82.66\%  \\
DOMFN  & Concat & 67.34\% & 73.72\% & 66.25\% & 72.44\% & 74.45\% & 68.57\%  & 83.56\%  & 82.62\%  \\
MSES   & Concat & 61.56\% & 66.83\% & 64.71\% & 70.63\% & 71.84\% & 66.55\%  & 84.18\%  & 83.60\%  \\\hline
PMR    & Average& 66.59\% & 70.30\% & 66.56\% & 71.93\% & 74.25\% & 68.60\%  & 83.60\%  & 82.49\%  \\
AGM    & Average& 67.07\% & 73.58\% & 66.02\% & 72.52\% & 74.83\% & 69.11\%  & 84.02\%  & 83.44\%  \\
MSLR   & Average& 65.46\% & 71.38\% & 65.91\% & 71.96\% & 72.52\% & 64.39\%  & 84.23\%  & \underline{83.69\%}  \\\hline
MLA*   &Weight  & 79.70\% & -       & 71.35\% & -       & -       & -        & -        & -        \\
MLA    &Weight  & 79.43\% & 85.72\% & 70.04\% & 74.13\% & 73.52\% & 67.13\%  & 84.26\%  & 83.48\%  \\
\hline 
\method&Average & \underline{79.84\%} & \underline{86.22}\% & \underline{72.28\%} & \underline{77.10\%} & {\bf 75.08\%} & {\bf 69.71\%}  & {\bf 84.72\%}  & {\bf 83.91\%}  \\
\method&Weight  & {\bf 80.38\%} & {\bf 86.45\%}   & {\bf 74.03\%} & {\bf 78.55\%}   & 74.89\% & \underline{69.17\%}    & {\bf 84.72\%}  & 83.32\%    \\
\Xcline{1-10}{0.7pt}
\end{tabular}
\end{table*}

\subsection{Experimental Settings}
\subsubsection{Baselines} To verify the effectiveness of our proposed method, we select a wide range of representative methods for comparison. They are OGR-GB~\cite{OGR-GB:conf/cvpr/WangTF20}, OGM~\cite{OGM:conf/cvpr/PengWD0H22}, DOMFN~\cite{DOMFN:conf/mm/0074ZGGZ22}, MSES~\cite{MSES:conf/acpr/FujimoriEKM19}, MSLR~\cite{MSLR:conf/acl/YaoM22}, Greedy~\cite{Greedy:conf/icml/WuJCG22}, MLA~\cite{MLA:journals/corr/abs-2311-10707}. Among these methods, OGR-GB, OGM, DOMFN, MSES are early fusion methods. MSLR and MLA are late fusion methods. And Greedy is a hybrid fusion method.

\subsubsection{Evaluation Protocols} Following the setting of OGM~\cite{DOMFN:conf/mm/0074ZGGZ22}, we use accuracy and mean average precision~(MAP) for \CREMAD~and \Kinetics~datasets. For \Twitter2015, \Sarcasm, and \NVGesture~datasets, we use accuracy and macro-F1 as evaluation metrics following the setting of the paper~\cite{Sarcasm:conf/acl/CaiCW19,Twitter15:conf/ijcai/Yu019}. The accuracy is used to measure the proportion of concordance between predictions and ground-truth labels. The MAP can be calculated by taking the mean of average precision for each category. And the macro-F1 can be calculated by averaging the F1 scores for each category.

\subsubsection{Implementation Details} Following the setting of OGM~\cite{OGM:conf/cvpr/PengWD0H22}, we use ResNet18~\cite{ResNet:conf/cvpr/HeZRS16} as the backbone to encode audio and video for \CREMAD~and \Kinetics~datasets. For \Twitter2015~and \Sarcasm~datasets, we adopt BERT~\cite{BERT:conf/naacl/DevlinCLT19} as the text encoder and ResNet50~\cite{ResNet:conf/cvpr/HeZRS16} as the image encoder following the setting of the paper~\cite{Twitter15:conf/ijcai/Yu019,Sarcasm:conf/acl/CaiCW19}. For \NVGesture~dataset, we follow the data preparation steps outlined in the paper~\cite{nvGesture:conf/icml/WuJCG22} and employ the I3D~\cite{I3D:conf/cvpr/CarreiraZ17} as uni-modal branches. For a fair comparison, all baselines also adopt the same backbone for the experiment. For \method, we explore a three-layer network, which can be denoted as ``$\text{FC}(Dim\times 256)\rightarrow\text{ReLU}\rightarrow\text{FC}(256\times 64)\rightarrow\text{FC}(64\times{c})$'', as classification head after features are extracted. Here, ``FC'' and ``ReLU'' denote the full-connected layer and ReLU~\cite{ResNet:conf/cvpr/HeZRS16} layer, respectively, and ``$Dim$'' denotes the dimension of features extracted by the encoder. For audio and video modalities, the dimension of the feature is 512. For image-text modalities and \NVGesture~dataset, the dimension is 1024. The gradient modification strategy is applied for the classification head for \method. Furthermore, for our proposed \method, we use SGD as the optimizer for the audio-video and \NVGesture~datasets, with a momentum of $0.9$ and weight decay of $1\times 10^{-4}$. The initial learning rate is set to be $1\times 10^{-2}$, and is divided by 10 when the loss is saturated. For image-text datasets~\cite{Twitter15:conf/ijcai/Yu019,Sarcasm:conf/acl/CaiCW19}, we use Adam as the optimizer, with an initial learning rate of $1\times 10^{-5}$. By using the cross-validation strategy with a validation set, the hyper-parameter scaling factor $\tau$ is set to be $0.4$ for all datasets. The hyper-parameter $\rho$ is set to be $1\times 10^{-15}$ and $1\times 10^{-10}$ for image/text modality and audio modality, respectively. During calculating cumulative variance, we set batch size as 12 for all datasets except \NVGesture. For \NVGesture~dataset, the batch size is set to 6 due to memory limitation. For all hyper-parameters, we utilize a cross-validation strategy on a validation set to determine their value. The experiments are performed with an NVIDIA RTX 3090 GPU.

\subsection{\small{Comparison with SOTA MML baselines}}\label{sec:main-exp}
To substantiate the superiority of our proposed method, we carry out comprehensive comparison experiments. 

\begin{table}[t]
\centering
\caption{Comparison with SOTA multimodal learning baselines on \NVGesture~dataset.}  
\label{tab:three-modalities-exp}
\small
\resizebox{0.5\textwidth}{!}{
\begin{tabular}{l|c|c|c|c}
\Xcline{1-5}{0.7pt}
Method   & Modality & Fusion & Accuracy       & Mac-F1 \\
\Xcline{1-5}{0.7pt}
      &    RGB   & N/A    & 78.22\%        &  78.33\% \\
Uni-Modal&     OF   & N/A    & 78.63\%        &  78.65\% \\
      &  Depth   & N/A    & 81.54\%        &  81.83\% \\
\hline
ORG-GB   &    All   & Concat & 82.99\%        &  83.05\% \\
MSES     &    All   & Concat & 81.12\%~($\downarrow$)  &  81.47\%~($\downarrow$) \\\hline
Baseline &    All   & Average& 82.57\%        &  82.68\% \\
MSLR     &    All   & Average& 82.37\%        &  82.39\% \\
AGM      &    All   & Average& 82.78\%        &  82.82\% \\\hline
MLA      &    All   & Weight & 83.73\%        &  83.87\% \\\hline
\method  &    All   & Average& {\bf 86.93\%}  &  {\bf 87.05\%} \\
\method  &    All   & Weight & {\bf 86.93\%}  &  \underline{87.03\%} \\
\Xcline{1-5}{0.7pt}
\end{tabular}
}
\end{table}

We first compare our method with SOTA multimodal learning baselines on \CREMAD, \Kinetics, \Twitter2015, and \Sarcasm~datasets. We report the accuracy, MAP, and Mac-F1 and the results are shown in Table~\ref{tab:main-exp}, where the best and the second-best results are shown in bold and underlining, respectively. Furthermore, the results of ``MLA*'' are referred from the origin paper of MLA. And the results of ``MLA'' are reproduced by us based on the authors' source code. For our proposed \method~with weighting fusion strategy version, we adopt the same weighting strategy as the MLA method for fair comparison. From Table~\ref{tab:main-exp}, we can observe that: (1). Compared with a wide range of SOTA baselines, our proposed method can achieve the best performance in all cases by substantially large margins. (2). \method~with a weighting fusion strategy can outperform \method~with an averaging fusion strategy on \CREMAD~and \Kinetics~datasets. While on \Twitter2015~and \Sarcasm~datasets, \method~adopting averaging fusion strategy can achieve comparable performance with \method~adopting weighting fusion strategy.


Then, we utilize \NVGesture~dataset for experiments to demonstrate the effectiveness of our method in scenarios involving more than two modalities. The results are reported in Table~\ref{tab:three-modalities-exp}, where the best and the second-best results are shown in bold and underlining, respectively. In Table~\ref{tab:three-modalities-exp}, Uni-Modal denotes that we utilize one modality to perform multimodal learning, which can be consider a basic baseline. And ``baseline'' denotes the basic multimodal learning algorithm without SAM-based optimization and gradient modification strategy. Surprisingly, we find that the performance of MESE is worse than that of Uni-Modal method, which is denoted by ``($\downarrow$)''. Furthermore, we can find that our method can achieve the best performance in scenarios involving three modalities.

\begin{table}[t]
\centering
\caption{Accuracy/MAP for ablation study on \Kinetics~dataset.}  
\label{tab:ablation-exp}
\resizebox{0.95\columnwidth}{!}{
\begin{tabular}{cc|c|c|c}
\Xcline{1-5}{0.7pt}
SAM  &GM & Audio & Video & Multi\\
\Xcline{1-5}{0.7pt}
\ding{56} & \ding{56} & 54.62\%/58.37\% & 53.12\%/56.69\% & 64.90\%/71.03\% \\\hline
\ding{52} & \ding{56} & 55.36\%/60.34\% & 54.05\%/59.31\% & 68.63\%/75.91\% \\\hline   
\ding{56} & \ding{52} & 55.35\%/58.43\% & 54.12\%/59.91\% & 70.01\%/76.12\% \\\hline
\ding{52} & \ding{52} & \bf{57.44\%}/\bf{60.43\%} & \bf{55.01\%}/\bf{60.69\%} & \bf{72.28\%}/\bf{77.10\%}\\
\Xcline{1-5}{0.7pt}
\end{tabular}
}
\end{table}

\subsection{Ablation Study}\label{sec:ablation}

\subsubsection{Effectiveness of SAM Loss and Gradient Modification} 

To fully explore the effectiveness of our proposed method, we study the influence of different components, including the SAM objective and gradient modification strategy. The results~(accuracy/MAP) of the ablation study on \Kinetics~dataset are shown in Table~\ref{tab:ablation-exp}, where ``SAM''/``GM'' denotes whether the SAM objective/gradient modification strategy is applied during training, respectively. And ``Audio'', ``Video'', and ``Multi'' denote that the accuracy and MAP are calculated based on the prediction of audio, video, and multiple modalities, respectively. From Table~\ref{tab:ablation-exp}, we can find that both SAM objective and gradient modification strategy can boost the performance in multimodal learning. 
\subsubsection{Necessity of Interactive Enhancement} 

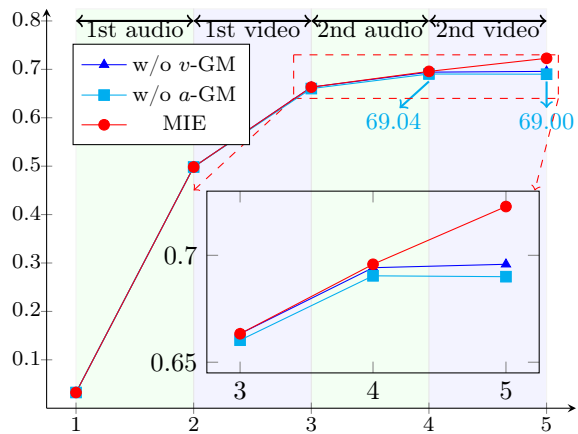
\begin{figure}[t]
\centering
\newsavebox{\mybox}
\savebox{\mybox}{
\begin{tikzpicture}
\begin{axis}[
xmin=0.75,
xmax=3.25,
ymin=0.645,
ymax=0.73,
ytick={0.65,0.7,0.75},
xtick={1,2,3},
xticklabels={3,4,5},
width=6cm,
height=4cm]
\addplot[mark=triangle*, color=blue, mark options={fill=blue}] coordinates {
(1, 0.6633)
(2, 0.6942)
(3, 0.6958)
};
\addplot[mark=square*, color=cyan, mark options={fill=cyan}] coordinates {
(1, 0.6603)
(2, 0.6904)
(3, 0.6900)
};
\addplot[mark=otimes*, color=red, mark options={fill=red}] coordinates {
(1, 0.6633)
(2, 0.6958)
(3, 0.7228)
};
\end{axis}
\end{tikzpicture}
}
\begin{tikzpicture}
\begin{axis}[
inner sep=0pt,
outer sep=0pt,
axis x line=center,
axis y line=center,
height={0.8\linewidth},
width={\linewidth},
xtick={1,2,...,5},
typeset ticklabels with strut,
enlarge x limits=false,
xticklabels={1,2,3,4,5},
ytick={0, 0.1,...,0.8},
xmin=0.75,
xmax=5.25,
ymin=0,
ymax=0.825,
legend style={at={(0.05,0.66)},anchor=south west},font=\footnotesize]
\filldraw [fill=green, opacity=0.05] (25,0)  rectangle (125,825); 
\filldraw [fill=blue,  opacity=0.05] (125,0) rectangle (225,825); 
\filldraw [fill=green, opacity=0.05] (225,0) rectangle (325,825); 
\filldraw [fill=blue,  opacity=0.05] (325,0) rectangle (425,825); 
\draw[->,thick, cyan] (325,675) -- (300,620) node [pos=1.2,below,font=\small] {69.04};
\draw[->,thick, cyan] (425,675) -- (425,620) node [pos=1.2,below,font=\small] {69.00};
\draw[<->,thick, black] ( 25,800) -- (125,800) node [pos=0.5,below,font=\small] {1st audio};
\draw[<->,thick, black] (125,800) -- (225,800) node [pos=0.5,below,font=\small] {1st video};
\draw[<->,thick, black] (225,800) -- (325,800) node [pos=0.5,below,font=\small] {2nd audio};
\draw[<->,thick, black] (325,800) -- (425,800) node [pos=0.5,below,font=\small] {2nd video};
\draw[-,dashed, red] (210,730) -- (210,640);
\draw[-,dashed, red] (435,730) -- (435,640);
\draw[-,dashed, red] (210,730) -- (435,730);
\draw[-,dashed, red] (210,640) -- (435,640);
\draw[->,dashed, red] (210,640) -- (125,450);
\draw[->,dashed, red] (435,640) -- (415,450);
\addplot+ [mark=triangle*, color=blue, mark options={fill=blue}] table {
1  0.0325
2  0.4983
3  0.6633
4  0.6942
5  0.6958
};\addlegendentry{w/o $v$-GM}
\addplot+ [mark=square*, color=cyan, mark options={fill=cyan}] table {
1  0.0325
2  0.4983
3  0.6603
4  0.6904
5  0.6900
};\addlegendentry{w/o $a$-GM}
\addplot+ [mark=otimes*, color=red, mark options={fill=red}] table {
1  0.0325
2  0.4983
3  0.6633
4  0.6958
5  0.7228
};\addlegendentry{\method}
\draw (axis cs: 3.25,.225)node{\usebox{\mybox}};
\end{axis}
\end{tikzpicture}
\caption{Interactive enhancement analysis~(best view in color).}
\label{fig:interactive}
\end{figure}

We carry out an experiment on \Kinetics~dataset to further analyze the necessity of interactive enhancement. The algorithm of our proposed \method~utilizes the well-learned model to modify the gradient of the model to be learned alternately. Due to the existence of dominant/non-dominant modality, we design two experiments for comparison on \Kinetics~dataset which contains audio modality~(dominant) and video modality~(non-dominant). For the first experiment, which is denoted as ``w/o $v$-GM'', we only utilize the model of audio modality to modify the gradient of video modality. On the contrary, the second experiment, which is denoted as ``w/o $a$-GM'', utilizes the model of video modality to modify the gradient of audio modality. According to Algorithm~\ref{algo:MAIE}, the training procedure can be divided into audio learning phase and video learning phase for \Kinetics~dataset. In the audio learning phase, the learned model of the video modality is used to modify the gradient of the audio modality. And the model of the audio modality is used to modify the gradient of the video modality during video learning phase. We report the accuracy at each phase during the training procedure in Figure~\ref{fig:interactive}. In Figure~\ref{fig:interactive}, we use blue and green colors to indicate the video and audio learning phase, respectively. 

From Figure~\ref{fig:interactive}, we can draw the following observations: (1). The overall performance of the interactive gradient modification is better than that of the non-interactive gradient modification, i.e., ``w/o $v$-GM'' and ``w/o $a$-GM''. (2). Compared with the ``w/o $a$-GM'', ``w/o $v$-GM'' can achieve better performance. In other words, the method which utilizes the model of the dominant modality to modify the gradient of the non-dominant is superior to the method which utilizes the model of the non-dominant modality to modify the gradient of the dominant modality. (3). Interestingly, we notice that during 4-th iteration~(video modality learning phase), there is a slight decrease in accuracy~(from 69.04\% to 69.00\%) for ``w/o $a$-GM'' method. That is to say when we shut down the gradient modification for video model training, the overall performance deteriorates even though we optimize the model for video modality.

\begin{figure}[t] 
\begin{minipage}[b]{0.485\linewidth}
\includegraphics[width=\linewidth]{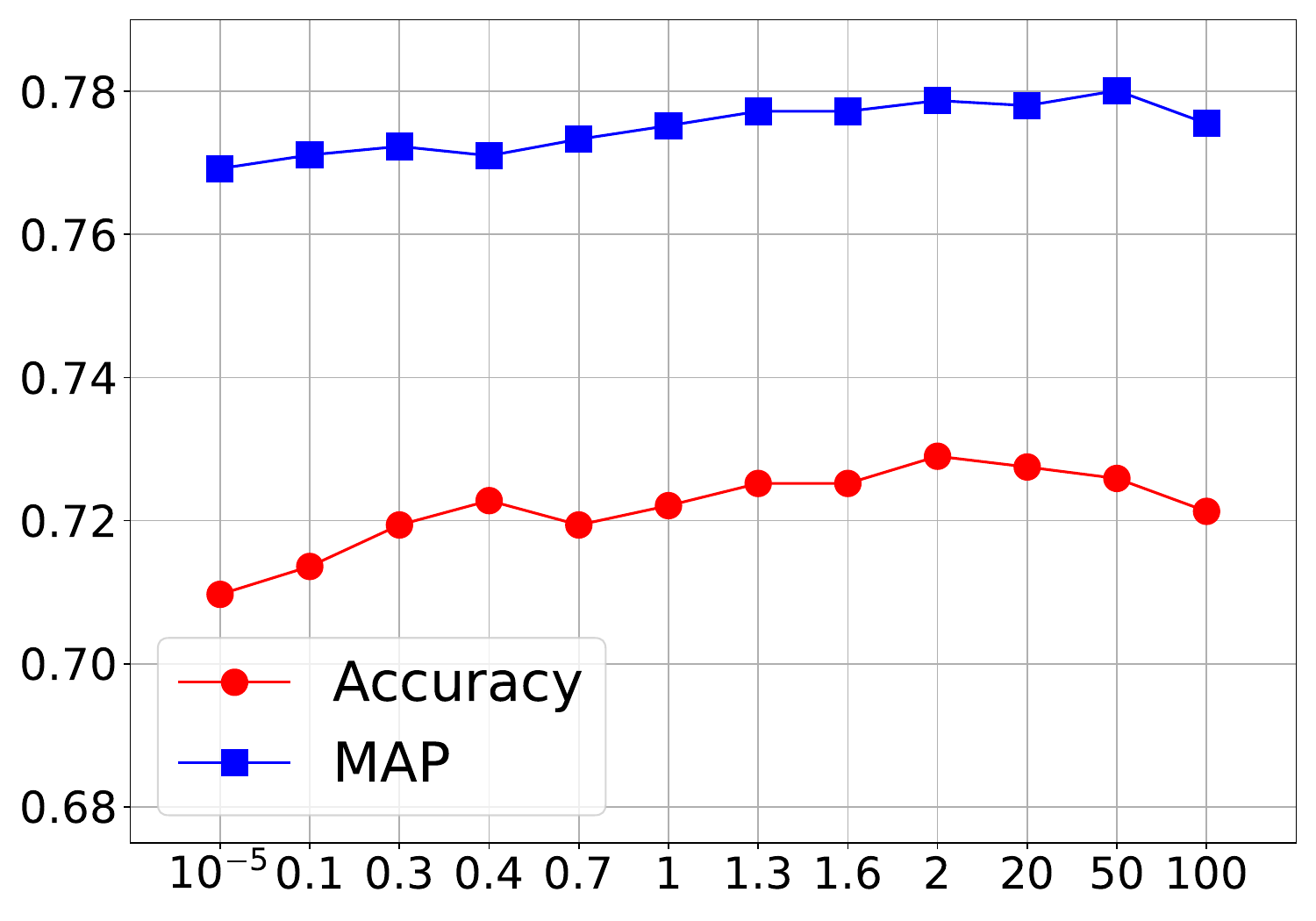}\\
\centering{(a). scaling factor $\tau$.}
\end{minipage} 
\begin{minipage}[b]{0.485\linewidth}
\includegraphics[width=\linewidth]{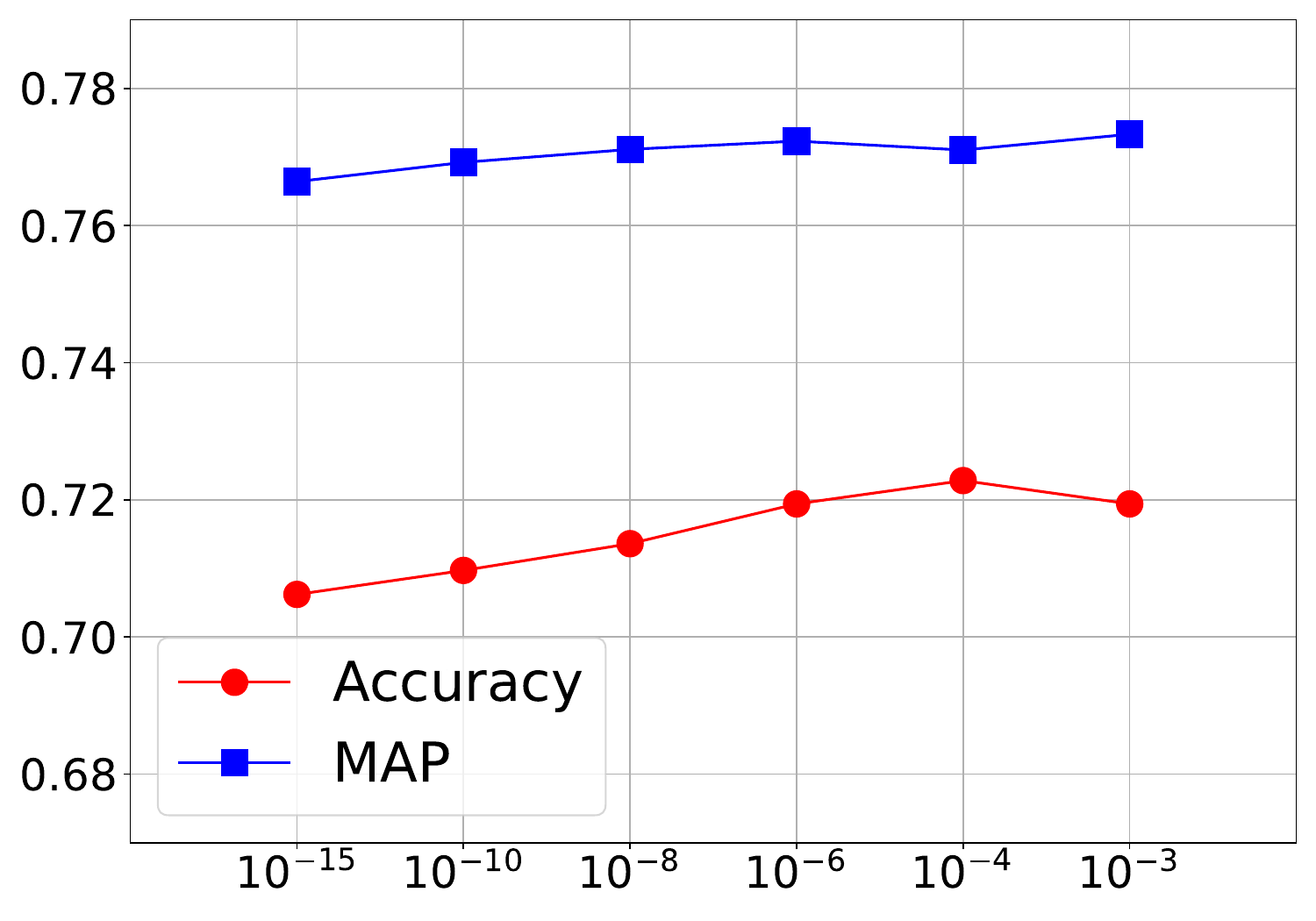}\\
\centering{(b). hyper-parameter $\rho$.}
\end{minipage} 
\caption{Sensitivity to $\tau$ and $\rho$.}\label{fig:sensitivity}
\end{figure}

\begin{figure*}[t] 
\begin{minipage}[b]{0.3\linewidth}
\includegraphics[width=.95\linewidth]{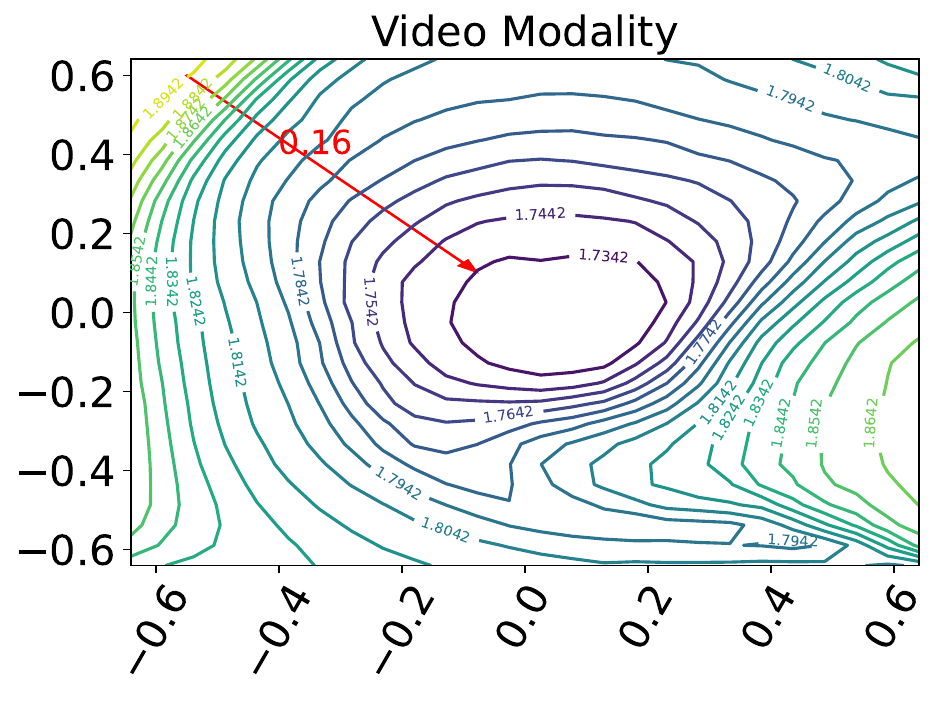}\\\centering{(a). Baseline} 
\end{minipage} 
\begin{minipage}[b]{0.3\linewidth}
\includegraphics[width=.95\linewidth]{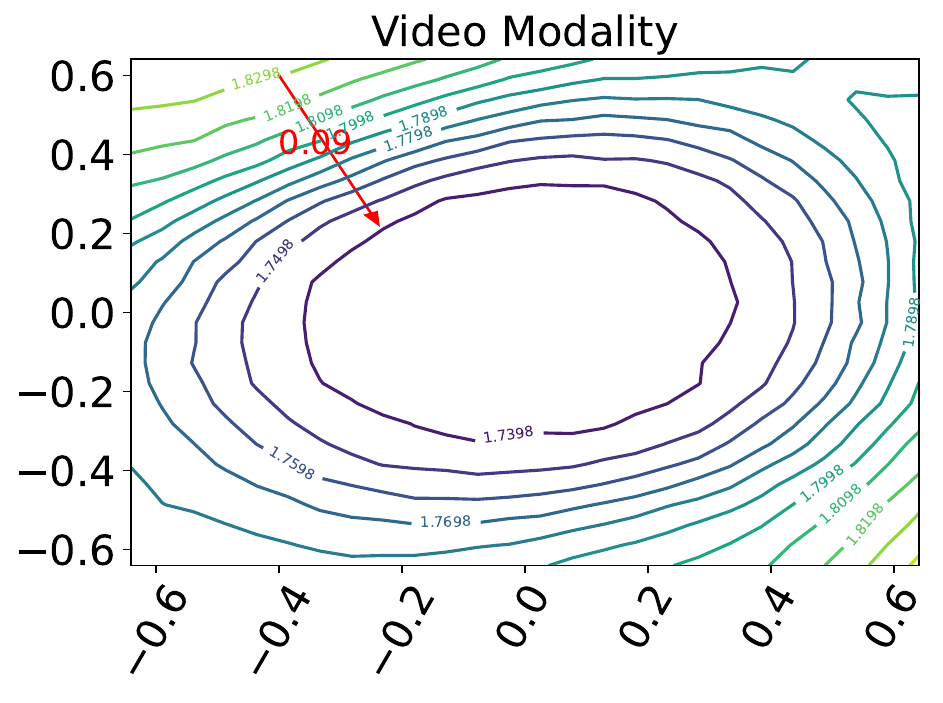}\\\centering{(b). MLA} 
\end{minipage} 
\begin{minipage}[b]{0.3\linewidth}
\includegraphics[width=.95\linewidth]{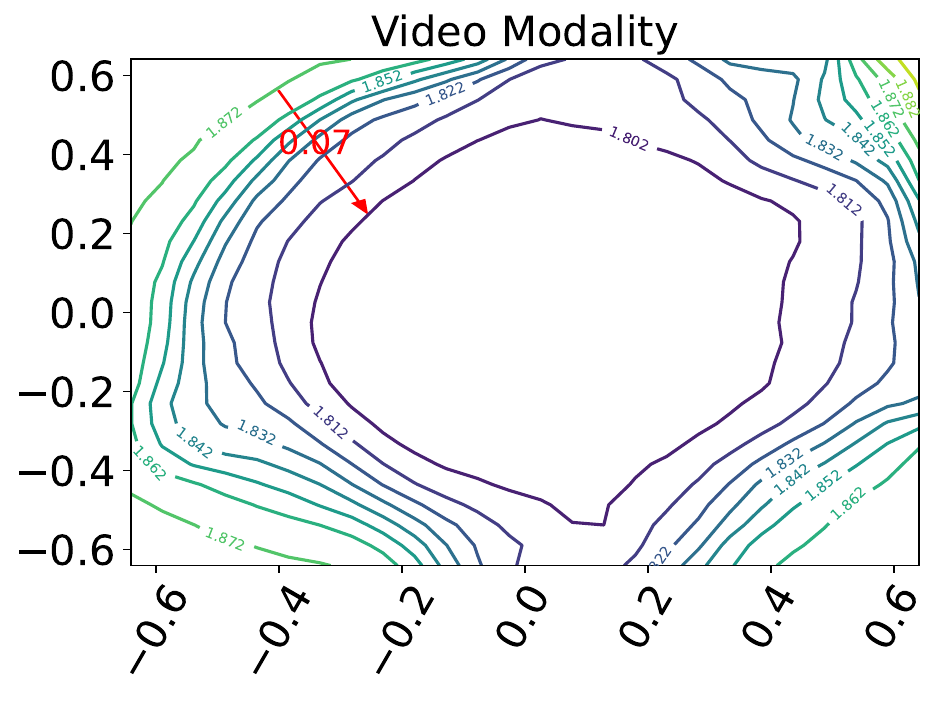}\\\centering{(c). \method} 
\end{minipage} \\
\begin{minipage}[b]{0.3\linewidth}
\includegraphics[width=.95\linewidth]{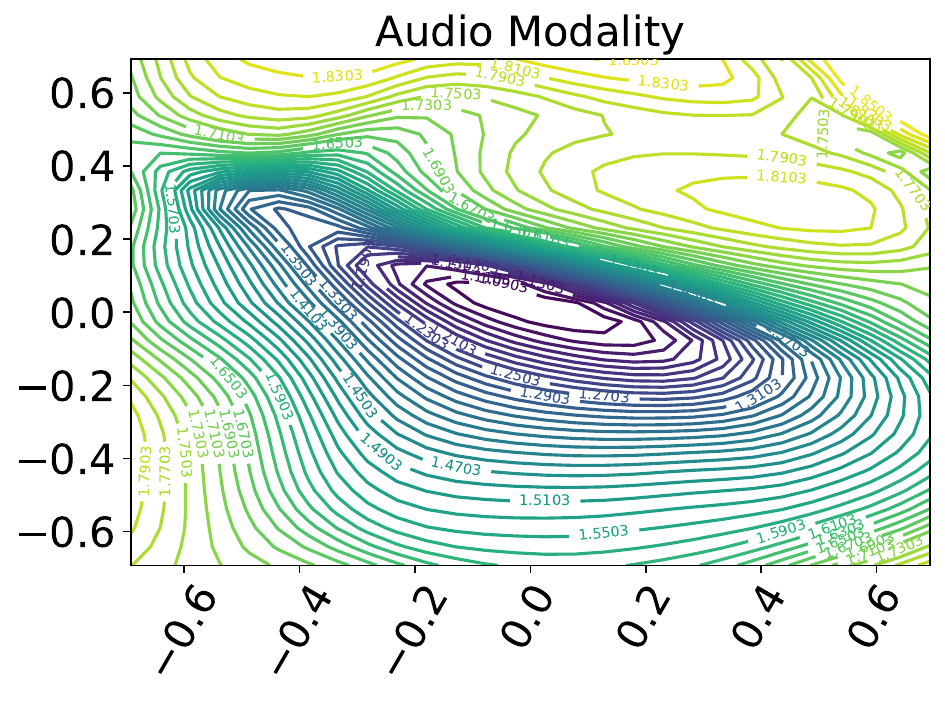} \\\centering{(d). Baseline} 
\end{minipage} 
\begin{minipage}[b]{0.3\linewidth}
\includegraphics[width=.95\linewidth]{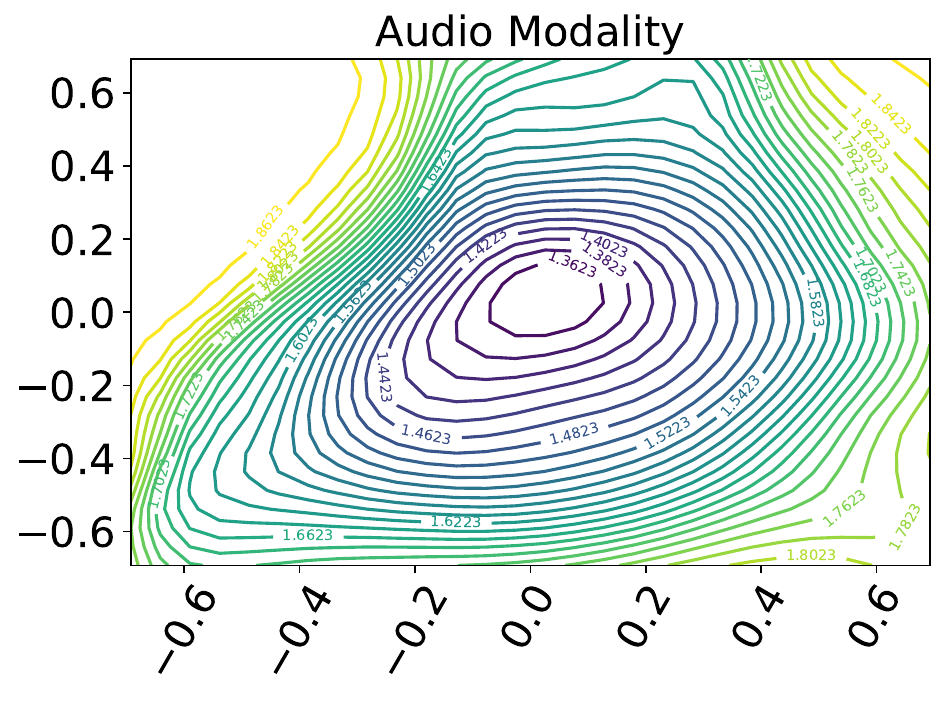}\\\centering{(e). MLA} 
\end{minipage} 
\begin{minipage}[b]{0.3\linewidth}
\includegraphics[width=.95\linewidth]{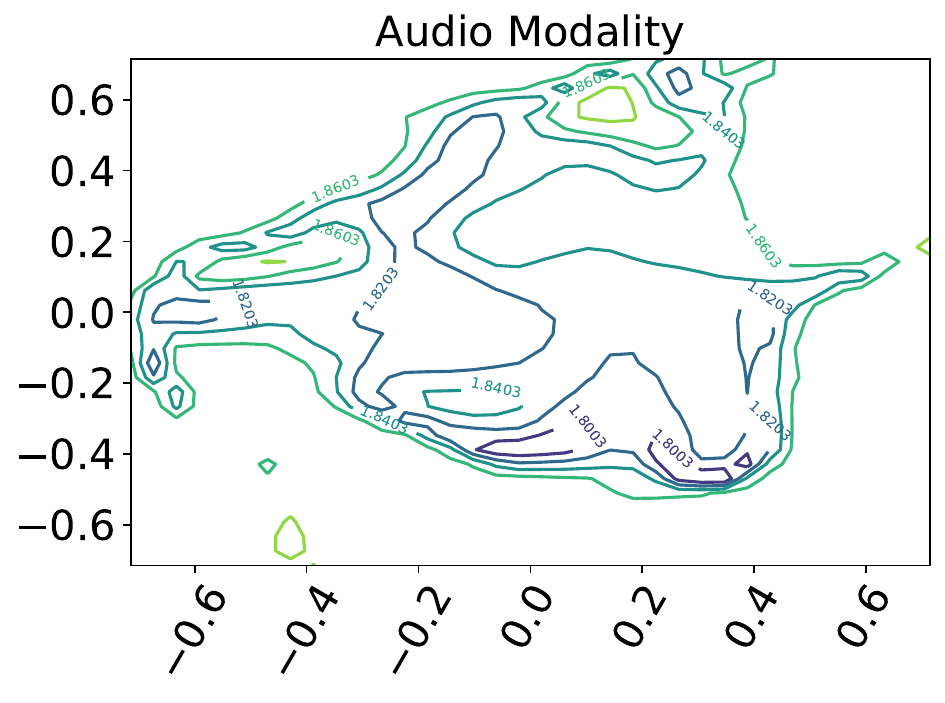}\\\centering{(f). \method} 
\end{minipage}
\caption{Loss landscape visualization. The first/second rows show the loss landscape for video/audio modalities, respectively.}
\label{fig:loss-visual} 
\end{figure*}

\subsection{Sensitivity to Hyper-Parameters}\label{sec:sensitity}
\subsubsection{Scaling Factor $\tau$} 

We study the influence of scaling factor $\tau$ on \Kinetics~dataset. We present the accuracy and MAP values with different $\tau$ from the range of $[10^{-5}, 100]$. The results are shown in Figure~\ref{fig:sensitivity}~(a). From Figure~\ref{fig:sensitivity}~(a), we can see that our proposed method is not sensitive to scaling factor $\tau$ in a large range. Furthermore, we can find that as the $\tau$ increases, the accuracy and MAP increase at the beginning and then remain unchanged. According to Equation~(\ref{eq:update-rule}) and the definition of $\Lambda^{(k)}$, the objective tends to perform learning along with flat directions when $\tau$ is enlarged, thus improving the performance of the model. 

\subsubsection{Hyper-Parameter $\rho$} 

We further study the influence of hyper-parameter $\rho$ on \Kinetics~dataset. We report the results with different $\rho$ in Figure~\ref{fig:sensitivity}~(b). From Figure~\ref{fig:sensitivity}~(b), we can find that \method~is not sensitive to hyper-parameter $\rho$ in a large range.

\subsubsection{The Scope of Gradient Modification} In this section, we study the influence of the scope of the gradient modification strategy. For the sake of simplicity, we carry out this experiment on \Kinetics~dataset, where the network architectures of audio and video modalities are the same, i.e., a ResNet18 as encoder and three full-connected layers as classification head. We select parameters along the deep to shallow layers of the neural network. And we define the scope of GM as the proportion of the selected network parameters to the total parameters. The results are shown in Table~\ref{tab:scope}, where ``0\%~(w/o GM)'' is used as the baseline and means that we don't perform gradient modification strategy during training. We can see that the best performance is achieved when we choose 30\% parameters for gradient modification. In contrast, choosing all parameters for gradient modification does not achieve the best performance. We argue that the essence of this phenomenon is that the shallow neural network focuses on the learning of visual feature patterns, and it is not suitable for too much perturbation, especially for heterogeneous data. Furthermore, we can also find that the performance of the method applying gradient modification strategy is better than that of the method which does not apply gradient modification.

\begin{table}[t]
\centering
\caption{Accuracy and MAP values with different scope of gradient modification.}  
\label{tab:scope}
\begin{tabular}{l|c|c}
\Xcline{1-3}{0.7pt}
Scope of GM                & Accuracy & MAP    \\
\Xcline{1-3}{0.7pt}
100\%                      & 72.13\% & 77.11\% \\
50\%                       & 72.21\% & 77.15\% \\
30\%                       & 72.56\% & 77.52\% \\
1.3\%~(Classification head)& 72.28\% & 77.10\% \\\hline
0\%~(w/o GM)               & 68.63\% & 75.91\% \\
\Xcline{1-3}{0.7pt}
\end{tabular}
\end{table}

\subsection{Further Analysis}
\subsubsection{Loss Landscape Visualization} To illustrate the impact of SAM optimization, we utilize the DNN visualization method~\cite{Visualization:conf/nips/Li0TSG18} to plot 2D loss function of baseline, MLA, and \method~on \Kinetics~dataset. Here, the baseline denotes the method we omit the SAM optimization and the gradient modification strategy. The results of the loss landscape are shown in Figure~\ref{fig:loss-visual}, where the first and second rows show the results for video and audio modality, respectively. The first, second, and third columns show the results of baseline, MLA, and \method, respectively. From Figure~\ref{fig:loss-visual}, we can find that the loss change of \method~is smaller than that of MLA and baseline. That is to say, the loss landscape of our method is flatter than that of baseline and MLA with the help of SAM optimization. Thus the generalization ability of \method~can be further improved.

\subsubsection{Magnitude of Singular Values} According to Equation~(\ref{eq:change}), the magnitude of singular values reflects the loss flatness of the direction indicated  by corresponding singular vectors. We report the singular values of different layers for the \method~and the method which does not adopt SAM loss~(denoted as ``\method~(w/o SAM)''). The results of \Kinetics~dataset are shown in Table~\ref{tab:singular-value}. In Table~\ref{tab:singular-value}, ``Max'' and ``Average'' denote the largest singular value and the average of singular values, respectively. From Table~\ref{tab:singular-value}, we can find that the singular values of \method~are smaller than that of the \method~(w/o SAM). In other words, the loss landscape of \method~is flatter than that of the method without SAM loss.
\begin{table}[t]
\centering
\caption{Magnitude of singular values.} 
\label{tab:singular-value}
\resizebox{0.45\textwidth}{!}{
\begin{tabular}{l|c|c|c}
\Xcline{1-4}{0.7pt}
Layer                                 & Stat.    & \method      & \method~(w/o SAM) \\\Xcline{1-4}{0.7pt}
{\multirow{2}{*}{FC($Dim\times 256$)}}& Max      & 3477.7       & 16926.6           \\\cline{2-4}
                                  & Average  & 7.2          & 33.7              \\\hline
{\multirow{2}{*}{FC($256\times 64$)}} & Max      & 3406.8       & 4549.7            \\\cline{2-4}
                                  & Average  & 15.8         & 19.9              \\\hline
{\multirow{2}{*}{FC($64\times c$)}}   & Max      & 288.3        & 373.0             \\\cline{2-4}
                                  & Average  & 35.7         & 19.5              \\
\Xcline{1-4}{0.7pt}
\end{tabular}
}
\end{table}

\begin{table}[t]
\centering
\caption{The Analysis of the Pretrained Model.}  
\label{tab:CLIP-verify}
\begin{tabular}{l|c|c|c}
\Xcline{1-4}{0.7pt}
Method      & Image    & Text   & Multi \\
\Xcline{1-4}{0.7pt}
CLIP        & 54.48\%  & 71.75\%& 72.52\%\\
CLIP+MLA    & 56.93\%  & 72.37\%& 72.95\%\\\hline
CLP+\method & {\bf 58.53\%}  & {\bf 73.19\%}& {\bf 73.58\%}\\
\Xcline{1-4}{0.7pt}
\end{tabular}
\end{table}

\subsubsection{Robustness of the Pretrained Model} We further explore the robustness of the large vision-language pre-trained model on \Twitter2015~dataset. Following the setting of MLA~\cite{MLA:journals/corr/abs-2311-10707}, we replace the backbones of image and text modalities as the corresponding encoders of CLIP~\cite{CLIP:conf/icml/RadfordKHRGASAM21}. We adopt the same three-layer network as the classification head for multimodal learning. Then we fine-tune the model on \Twitter2015~dataset. We report accuracy results in Table~\ref{tab:CLIP-verify}, where ``CLIP+MLA'' and ``CLIP+\method'' denote that during fine-tuning we apply MLA and \method, respectively. From Table~\ref{tab:CLIP-verify}, we can find that: (1). MLA and \method~can achieve better performance compared with CLIP. (2). Our proposed method can boost higher improvement based on CLIP encoder compared with MLA.

\section{Conclusion}\label{sec:conclusion}
In this paper, we propose a novel multimodal learning method, called modal-aware interactive enhancement~(\method), for multimodal learning. By introducing the SAM-based optimization, we design a modal-aware SAM objective to smooth the learning objective during the forward phase. Furthermore, we propose a gradient modification strategy with the help of the geometry property of SAM. In this way, we can improve the generalization ability and alleviate the modality forgetting phenomenon simultaneously, leading to better performance in multimodal learning tasks. Extensive experiments demonstrate the superiority of \method~across five widely used datasets.

\bibliographystyle{plain}
\bibliography{main}

\appendix
\begin{center}
\Large\bf{Appendix}
\end{center}

\section{Additional Experimental Results}
\subsubsection{Loss Landscape Visualization on \Twitter2015~dataset} 
\begin{figure}[t]
\begin{minipage}[b]{0.45\linewidth}
\includegraphics[width=.975\linewidth]{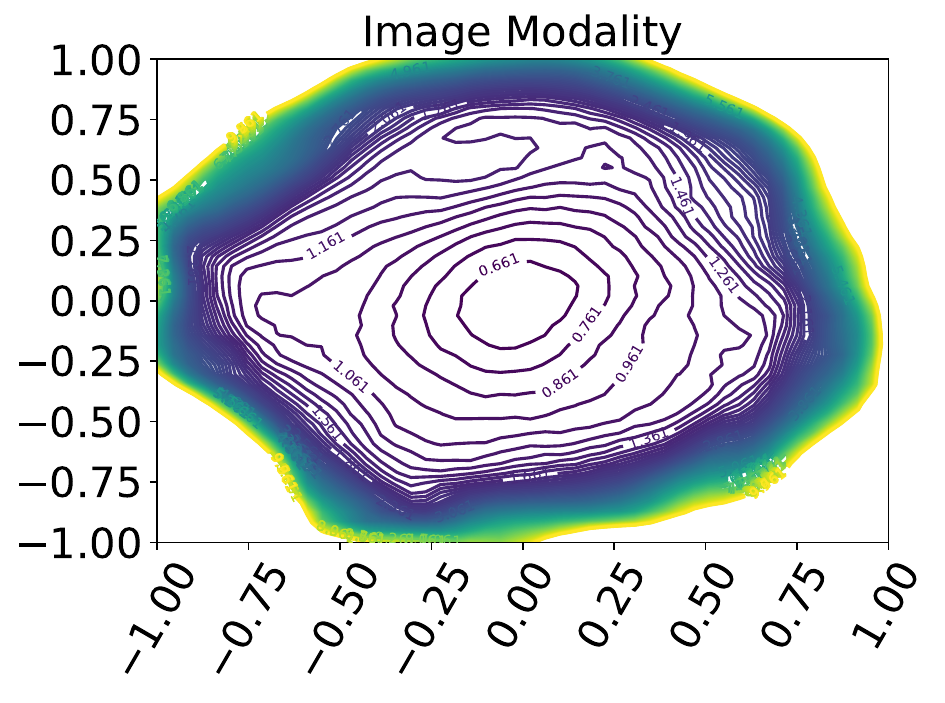} \\\centering{(a). Baseline} 
\end{minipage} 
\begin{minipage}[b]{0.45\linewidth}
\includegraphics[width=.975\linewidth]{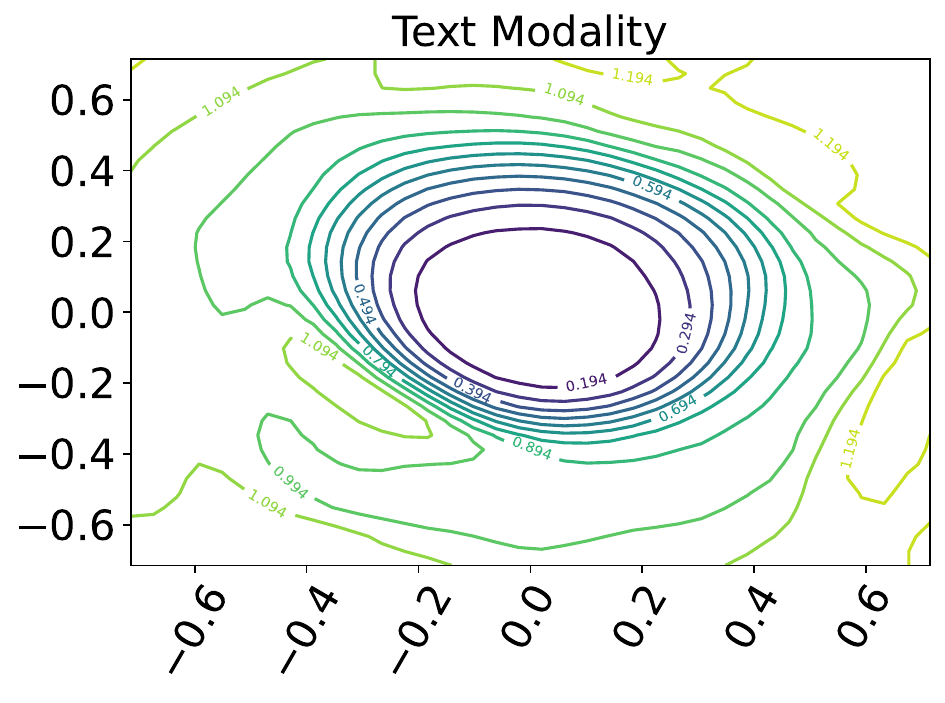}
\\\centering{(b). Baseline} 
\end{minipage} \\
\begin{minipage}[b]{0.45\linewidth}
\includegraphics[width=.975\linewidth]{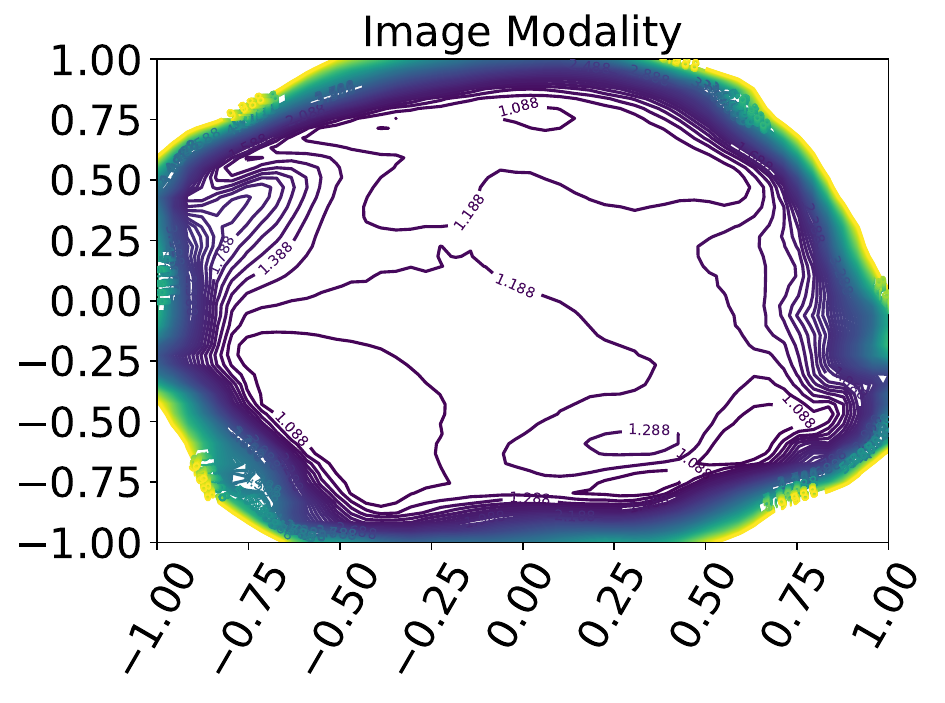} 
\\\centering{(c). MLA} 
\end{minipage} 
\begin{minipage}[b]{0.45\linewidth}
\includegraphics[width=.975\linewidth]{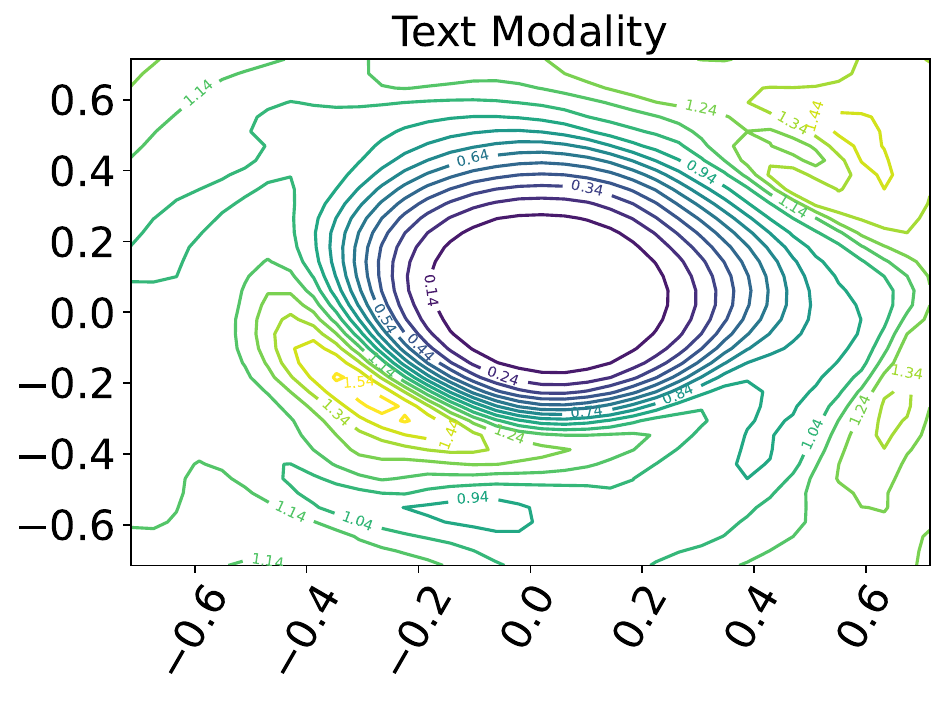}
\\\centering{(d). MLA} 
\end{minipage} \\
\begin{minipage}[b]{0.45\linewidth}
\includegraphics[width=.975\linewidth]{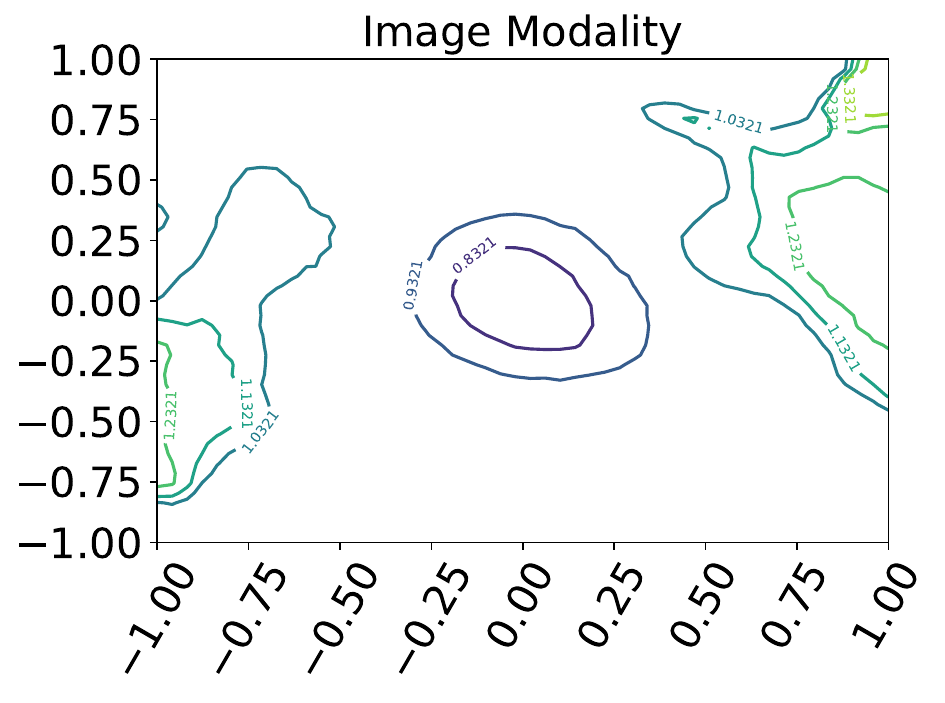} 
\\\centering{(e). \method} 
\end{minipage} 
\begin{minipage}[b]{0.45\linewidth}
\includegraphics[width=.975\linewidth]{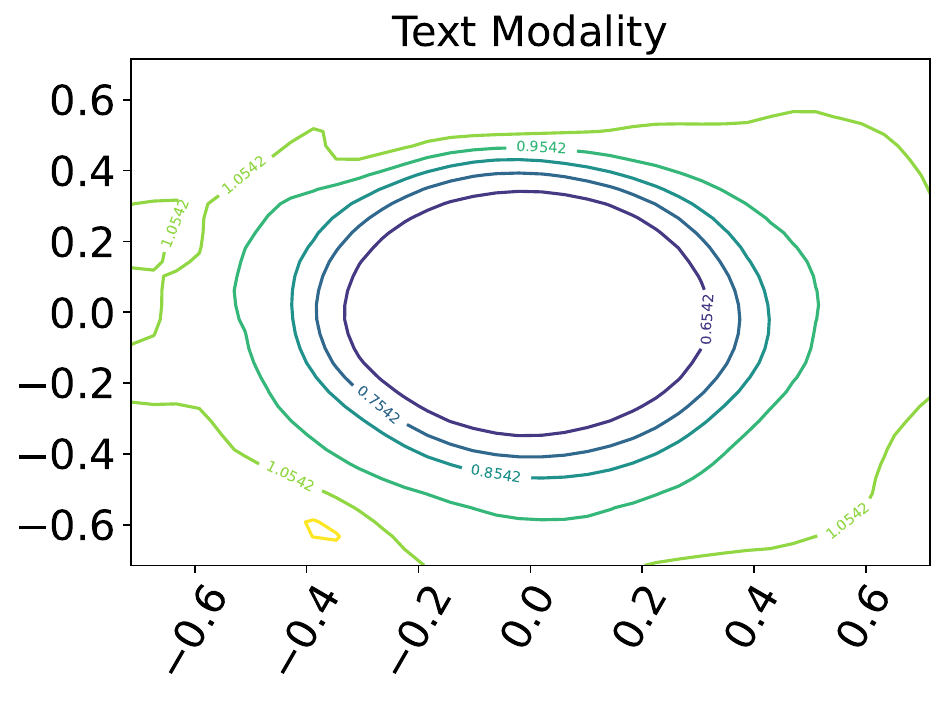}
\\\centering{(f). \method} 
\end{minipage} 
\caption{Loss landscape visualization on \Twitter2015~dataset.}
\label{fig:loss-visual-twitter} 
\end{figure}

We report the loss landscape visualization results on \Twitter2015~dataset in Figure~\ref{fig:loss-visual-twitter}. In Figure~\ref{fig:loss-visual-twitter}, the first and second columns denote the loss landscape of image and text modalities, respectively and the first, second, third rows denote the loss landscape of Baseline, MLA, and \method, respectively. From Figure~\ref{fig:loss-visual-twitter}, we can find that the loss landscape of \method~is flatter than that of MLA and baseline. Hence, our method can generalize better than MLA and baseline.

\subsubsection{Further Analysis of Singular Values}
Noticing that we does not adopt normalization operation when we calculate cumulative variance matrix, we report the evaluation criteria and singular values of \method~and normalized-\method. We report the results in Table~\ref{tab:singular-value-further}. From Table~\ref{tab:singular-value-further}, we can find that although the absolute values of singular values are different, the accuracy and MAP are the same. The reason is that we adopt normalization when we define matrix $\Sigma$.
\begin{table}[t]
\centering
\caption{Magnitude of singular values.} 
\label{tab:singular-value-further}
\begin{tabular}{l|c|c|c}
\Xcline{1-4}{0.7pt}
Layer                                 & Stat.    & \method      & Normalized-\method \\\Xcline{1-4}{0.7pt}
{\multirow{2}{*}{FC($Dim\times 256$)}}& Max      &    3477.7    &    0.1299         \\\cline{2-4}
                                  & Average  &    7.2       & 0.0003 \\\hline
{\multirow{2}{*}{FC($256\times 64$)}} & Max      &    3406.8    &     0.1272        \\\cline{2-4}
                                  & Average  &    15.8      & 
0.0006         \\\hline
{\multirow{2}{*}{FC($64\times c$)}}   & Max      &  288.3       &    0.0109         \\\cline{2-4}
                                  & Average  &  35.7        &    0.0013         \\\hline
\multicolumn{2}{c|}{{Accuracy}}                  &  72.28\%     &     72.28\%       \\\hline   
\multicolumn{2}{c|}{{MAP}}                       &  77.10\%     &     77.10\%       \\
\Xcline{1-4}{0.7pt}
\end{tabular}
\end{table}

\section{Notations}
We summarize the notation we used in this paper in Table~\ref{tab:notation}.
\begin{table}[t]
\centering
\caption{Notation summary.} 
\label{tab:notation}
\begin{tabular}{l|c}
\Xcline{1-2}{0.7pt}
Notation                                    & Description                        \\\Xcline{1-2}{0.7pt}
$n$                                         & The number of training data        \\\hline
$m$                                         & The number of modality             \\\hline
$c$                                         & The number of category             \\\hline
$\x_i^{(k)}$                                & $k$-th modality $x$ for $i$-th data\\\hline
$\y_i$                                      & Category label for $i$-th data     \\\hline
$\DM=\{\XM^{(k)}\}_{k=1}^m$                 & Training set                       \\\hline
$\XM^{(k)}=\{\x^{(k)}_1,\cdots,\x^{(k)}_n\}$& Data points of $k$-th modality     \\\hline
$\p^{(k)}_i$                                & Prediction of $i$-th data point    \\\hline
$\ell(\cdot)$                               & Cross entropy loss                 \\\hline
$L(\cdot)$                                  & Empirical loss function            \\\hline
$f(\cdot)$                                  & Fusion function                    \\\hline
$L^{\text{SAM}}(\cdot)$                     & SAM loss function                  \\\hline
$\epsilon$                                  & Perturbation                       \\\hline
$\rho$                                      & Perturbation magnitude restriction \\\hline
$\Theta$                                    & Parameter space                    \\\hline
$\z_i^{(j)}$                                & Feature extracted by encoder       \\\hline
$\Y_t^{(k)}$                                & Covariance of $t$-th batch         \\\hline
$\bar\Y_t^{(k)}$                            & Cumulative variance                \\\hline
$\U^{(k)}$                                  & Left-singular vectors              \\\hline
$\V^{(k)}$                                  & Right-singular vectors             \\\hline
$\Lambda^{(k)}$                             & Singular values                    \\\hline
$\Sigma^{(k)}$                              & Coefficient matrix                 \\\hline
$\T^{(k)}$                                  & Gradient modification matrix       \\\hline
$\tau$                                      & Scaling factor                     \\\hline
$\eta^{(l)}$                                & Learning rate                      \\
\Xcline{1-2}{0.7pt}
\end{tabular} 
\end{table}

\end{document}